%% file: acl_latex.tex
\pdfoutput=1

\documentclass[11pt]{article}

\usepackage[final]{acl}

\usepackage{times}
\usepackage{latexsym}

\usepackage[T1]{fontenc}

\usepackage[utf8]{inputenc}

\usepackage{microtype}

\usepackage{inconsolata}

\usepackage{graphicx}
\usepackage{amsmath}
\usepackage{multirow}
\usepackage{subcaption}
\usepackage{float}
\usepackage{algorithm}
\usepackage{algcompatible}
\usepackage{xspace}
\usepackage{authblk}
\usepackage[most]{tcolorbox}

\newtcolorbox{prompt}[1][]{%
  sharp corners,
  enhanced,
  colback=white,
  height=9cm,
  attach title to upper,
  #1
}

\newcommand{\metricName}{SMuDGE\xspace} 

\title{Where is this coming from? Making groundedness count in the evaluation of Document VQA models}

\author[1,2]{\textbf{Armineh Nourbakhsh}}
\author[1]{\textbf{Siddharth Parekh}}
\author[2]{\textbf{Pranav Shetty}}
\author[1]{\textbf{Zhao Jin}}
\author[2]{\authorcr\textbf{Sameena Shah}}
\author[1]{\textbf{Carolyn P. Ros\'{e}}}
\affil[1]{Language Technologies Institute, Carnegie Mellon University}
\affil[2]{J.P. Morgan, New York}
\affil[ ]{\texttt{anourbak@cs.cmu.edu}}

\begin{document}
\maketitle
\begin{abstract}
Document Visual Question Answering (VQA) models have evolved at an impressive rate over the past few years, coming close to or matching human performance on some benchmarks. We argue that common evaluation metrics used by popular benchmarks do not account for the semantic and multimodal groundedness of a model's outputs. As a result, hallucinations and major semantic errors are treated the same way as well-grounded outputs, and the evaluation scores do not reflect the reasoning capabilities of the model. In response, we propose a new evaluation methodology that accounts for the groundedness of predictions with regard to the semantic characteristics of the output as well as the multimodal placement of the output within the input document. Our proposed methodology is parameterized in such a way that users can configure the score according to their preferences. We validate our scoring methodology using human judgment and show its potential impact on existing popular leaderboards. Through extensive analyses, we demonstrate that our proposed method produces scores that are a better indicator of a model's robustness and tends to give higher rewards to better-calibrated answers. 
\end{abstract}

\section{Introduction}
Visual Question Answering (VQA) over multimodal documents requires joint reasoning over textual, spatial, and visual signals. Several benchmarks have been proposed to measure the performance of SotA models on this task, including single-page and multi-page VQA \cite{docvqa2021, mpdocvqa2023, infographicvqa2022, dude2023, doccvqa2021}. In these benchmarks, the ground truth answer is expressed as a sequence of tokens and evaluated against the sequence of tokens produced by each model. As such, the evaluation metrics used by these benchmarks focus on the surface similarity between the model output and the ground-truth answer. This misses two key aspects of the model's output: 1) Is it aligned with the expected semantic category? For example, if the ground truth is a number, is the model also producing a number? 
2) Can it be located within the input document? In other words, is the model hallucinating a response or is it generating something based on the document (even if it is wrong)? Grounded responses help determine the provenance of the model output and verify its accuracy. 

\input{latex/figexample}
\input{latex/tabexample}

Figure \ref{fig:snippets} and Table \ref{tab:example} illustrate this using an example from the DocVQA benchmark \cite{docvqa2021}, which uses Normalized Levenshtein Distance as its evaluation metric \cite{nls1966}. Given the two excerpts from an image document in Figure \ref{fig:snippets}, two questions are listed in Table \ref{tab:example}. The first question, ``How many mgs of iron is in enriched farina?'', requires the model to reason over a tabular structure and produce the answer ``12''. If the model produces ``26'' as the answer, it will be rewarded by a score of 0.5 because ``26'' shares one digit with the ground-truth answer, ``12''. In contrast, if the model produces ``8.5'' as the answer, it will not be rewarded, as there is no overlap with the ground-truth. This is potentially problematic, as the first answer is not mentioned anywhere on the page, and can therefore be considered hallucinatory. The second answer, although inaccurate, captures a number that is present in the table in Figure \ref{fig:snippetstab}, and is located on the same column as the ground-truth, potentially signifying some level of tabular reasoning by the model. A more robust evaluation metric would provide a small reward to the second answer, and give the first answer a score of 0.0.

Another question, ``How much added iron do premodified infant formulas contain?'', requires verbal reasoning over the paragraph in Figure \ref{fig:snippetspar}. If a model responds by ``up to 12 mgs'', it is penalized for its surface dissimilarity to the ground-truth answer, ``up to 12 milligrams''. In contrast, if the model produces ``up to 1z milligrams'', it is awarded a higher score because its answer has a larger overlap with the ground-truth. Again, this is problematic, as the second answer misrecognizes a key component of the ground truth (i.e. the number) and as such indicates a completely inaccurate quantity. A more robust evaluation metric should reward a higher score for the first answer than for the second.

In this paper, we propose a new evaluation methodology, which we name Semantics and MUltimodal Document Grounded Evaluation (\metricName). \metricName addresses the above issues by grounding the similarity score in the expected output type (i.e. numeric, textual, or hybrid answers). We also add a new component---a multimodal grounding score that determines whether the model's output is located within the input document, and where it is located in relation to the ground-truth. In the Document AI literature, this form of multimodal grounding is also referred to as localization \cite{localization2015}. 

As \citet{position2024} argued, grounding is an important requirement (and challenge) for the operationalization of Document VQA models especially in enterprise domains. Nevertheless it is difficult to determine how much grounding might matter to one downstream application versus another. Therefore, we design our evaluation approach to accommodate different settings by allowing users to set the preferred weights for each component.

Concretely, our study makes the following contributions to the field:
\begin{enumerate}
\item We propose a new evaluation framework (\metricName) that accounts for the groundedness of outputs and the semantic type of the output. We design \metricName to be configurable and easy to tune for downstream applications.
\item Using \metricName, we re-evaluate the performance of SotA models on four common Document VQA benchmarks, and analyze the impact of grounding on the ranking of each leaderboard.
\item We perform a detailed analysis of the types of questions and answers most impacted by grounding-sensitive criteria, and propose a configurable setting that allows the downstream users of each model to tune the evaluation to their needs.
\item Our analyses show that \metricName produces scores better aligned with human preferences.  
\item We experimentally demonstrate that better-grounded generation is associated with better calibrated outputs.
\item Lastly, our analyses show that \metricName rewards models that are more robust to variations in tasks and datasets.
\end{enumerate}

\section{Background}
In recent years, generative multimodal models have made major strides in Visual Question Answering over image documents. As an example, as of January 2025, the top-performing model on the DocVQA leaderboard is within 2 points of human performance\footnote{\url{https://rrc.cvc.uab.es/?ch=17&com=evaluation&task=1}}. 

A key challenge of generative models is that their output is difficult to ground within the input document \cite{k2q2024} (this is also known as the challenge of localization \cite{localization2015}). Localization is a common requirement in many real-world applications, especially in enterprise domains where maintaining a proper lineage of data is crucial from a governance perspective \cite{position2024}. Since generative models produce sequences that are sampled from their vocabulary, they are not guaranteed to generate answers that are based on the input, unless forced to do so via grounded decoding (e.g. as in \cite{grounddec2024}). This in turn makes it difficult to detect hallucinations, establish the provenance of the model's generations, or measure the reliability of its outputs, all of which limit the applicability of such models in many enterprise domains \cite{position2024}. 

This problem is compounded by the fact that most popular Document VQA benchmarks do not account for grounding in their evaluation criteria. A common metric used by these benchmarks is Average Normalized Levenshtein Similarity (ANLS), as proposed by \citet{docvqa2021}, which measures the similarity between the ground truth and predicted answers based on their edit distance. As an example, the words `apple' and `app1e' have a Levenshtein distance of $1$, a normalized Levenshtein distance of $0.2$, and an NLS of $1-0.2=0.8$. If the score for a ground-truth/prediction pair is below a predefined threshold (typically set to 0.5 \cite{scene2019, mpdocvqa2023, infographicvqa2022, anls*2024}), the score is flattened to zero, otherwise the raw similarity score is used. 

The flexibility that the NLS metric provides allows the benchmarks to handle minor errors such as character misspellings resulting from poor Optical Character Recognition, without over-penalizing the models. Contrast this with a metric that relies on n-gram overlap metrics, or cosine similarity of distributed representations. Such metrics might consider ``\texttt{apple}'' and ``\texttt{app1e}'' to be very dissimilar words, given a single-character difference between them (where the the letter ``\texttt{l}'' has been replaced by the digit ``\texttt{1}''). Nevertheless, relying solely on surface similarity carries other risks for robust evaluation:
1) Surface similarity does not account for how a small change in the characterization of an answer can impact its meaning (e.g. changing a single digit in a number can change its value by a large magnitude). 2) Surface similarity cannot distinguish between answers that can be traced back to the input document, and those that can result from hallucination.

More recently, some studies have noted the shortcomings of common evaluation metrics in the field of multimodal document understanding, and proposed alternatives \cite{treeform2024}. Most notably, \citet{anls*2024} proposed ANLS*, a data-type-aware metric that can be used for single or multi-piece extraction and QA over documents. While it addresses many challenges of the ANLS metric, ANLS* is not designed to capture the multimodal groundedness of model outputs. 
In contrast, we focus on the challenge of measuring groundedness for extractive VQA over documents, where correct answers are guaranteed to be expressed in the input. We propose a configurable evaluation method that not only accounts for the groundedness of predictions, but also incorporates the semantic type of the output, similar to ANLS*. To the best of our knowledge, this is the first study that examines the impact of groundedness in evaluating Document VQA models. The following section describes our proposed approach in detail.


\section{Proposed methodology}
To measure the impact of groundedness in Document VQA performance, we develop a composite score to rate the output of each model. To ensure that the score can be applied to all models and benchmarks, we assume access to four objects only: 1) The question. 2) The ground truth answer. 3) The answer provided by the model. 4) A dictionary of words and corresponding bounding box coordinates extracted from the input document. This dictionary can be obtained by applying any OCR tool to the document, though the quality of character recognition often differs between different providers. Most benchmarks provide this dictionary as part of their data release.

In the next two subsections, we describe how we calculate two subscores: 1) The multimodal grounding score addresses the question of whether the predicted answer can be located within the input document, and if so, where it is located with respect to the ground truth answer. 2) The type-aware surface similarity score evaluates the predicted answer based on its type, i.e. numeric, textual, or hybrid.

\subsection{Multimodal groundedness}
\label{sec:grounding}
Given a question $q_i$, a ground truth answer $t_i$, and a predicted answer $a_i$, we develop a score $g_i$ that places $a_i$ within the originating document (composed of words $w_1, w_2, \cdots, w_N$ and corresponding bounding boxes $b_1, b_2, \cdots, b_N$) and measures its distance to $t_i$. We do this in two steps:

\textbf{Locating the predicted answer ($a_i$) and the ground truth ($t_i$) within the document.} To locate $t_i$ within the document, we find a continuous sequence of words $w_k, w_{k+1}, \cdots, w_{k+n}$ that matches $t_i$.\footnote{Note that a multimodal document is a 2-D artifact, and therefore a ``continuous sequence'' can extend in multiple directions, depending on the reading order of the page. Most commercial OCR packages such as \texttt{Textract} segment each page based on semantic information, e.g. an address block is presented as one segment, even if it contains multiple lines. We therefore rely on the segments provided by these packages to determine continuity. In the absence of such information, a graph representation of the document can be used as a proxy. In Appendix \ref{app:grounding}, we provide an algorithm that can be used to ground the sequence using this graph representation.} If no such segment is found (say, due to OCR errors), then we find a sequence that has the highest Normalized Levenshtein Similarity (NLS) to $t_i$. We name this sequence $\mathbf{w}_{t_i}$ and the corresponding bounding box $\mathbf{b}_{t_i}$, which is calculated by merging $b_k, b_{k+1}, \cdots, b_{k+n}$\footnote{See Appendix \ref{app:bboxes} for additional details.}. Similarly, we find the sequence $\mathbf{w}_{a_i}$ and the corresponding bounding box $\mathbf{b}_{a_i}$ by placing $a_i$ within the document. Note that $a_i$ is not guaranteed to be found on the page, for instance in case of hallucinations. If we can't find a $\mathbf{w}_{a_i}$ such that $\text{NLS}(a_i, \text{concat}(\mathbf{w}_{a_i})) > 0.3$\footnote{See Appendix \ref{app:threshold} for more information on how this threshold was selected.}, then we define $\mathbf{b}_{a_i}$ as:
\begin{align}
\label{eq:hallu}
    & \resizebox{0.9\columnwidth}{!}{%
    $[\mathbf{b}_{t_i}^\texttt{left}, \mathbf{b}_{t_i}^\texttt{top}, -\texttt{width}_i-\mathbf{b}_{t_i}^\texttt{right}, -\texttt{height}_i-\mathbf{b}_{t_i}^\texttt{bottom}]$}
\end{align}

where $\mathbf{b}_.^{\texttt{left}}, \mathbf{b}_.^{\texttt{top}}, \mathbf{b}_.^{\texttt{right}}, \mathbf{b}_.^{\texttt{bottom}}$ indicate the four coordinates of the bounding box $\mathbf{b}_.$ and $\texttt{width}_i, \texttt{height}_i$ indicate the width and height of the page, respectively. In other words, we use the bounding box of the ground-truth answer $t_i$ and mirror its bottom right corner in the negative space. This ensures that the distance between $\mathbf{b}_{t_i}$ and $\mathbf{b}_{a_i}$ is measured as 1 (see below).

\textbf{Measuring the distance.} Next, we measure $d_i$, the distance between $\mathbf{b}_{a_i}$ and $\mathbf{b}_{t_i}$. We do this by first finding the centroid of each bounding box, and then measuring the Normalized Manhattan Distance (NMD) between the centroids.\footnote{See Appendix \ref{app:norm} for a discussion of alternative distance normalization methods.} In other words:
\begin{align}
\label{eq:nmd}
    d_i = & \\ \notag
    & \resizebox{0.85\columnwidth}{!}{%
    $|\frac{\mathbf{b}_{t_i}^{\texttt{right}}}{2 \times \texttt{width}_i} - \frac{\mathbf{b}_{t_i}^{\texttt{left}}}{2 \times \texttt{width}_i} - \frac{\mathbf{b}_{a_i}^{\texttt{right}}}{2 \times \texttt{width}_i}  + \frac{\mathbf{b}_{a_i}^{\texttt{left}}}{2 \times \texttt{width}_i}| + $} \\ \notag
    & \resizebox{0.83\columnwidth}{!}{%
    $|\frac{\mathbf{b}_{t_i}^{\texttt{bottom}}}{2 \times \texttt{height}_i} - \frac{\mathbf{b}_{t_i}^{\texttt{top}}}{2 \times \texttt{height}_i} - \frac{\mathbf{b}_{a_i}^{\texttt{bottom}}}{2 \times \texttt{height}_i}  + \frac{\mathbf{b}_{a_i}^{\texttt{top}}}{2 \times \texttt{height}_i}| $ } \notag
\end{align}

If the predicted answer $a_i$ cannot be located within the document, the formulation presented in Equation \ref{eq:hallu} yields $d_i = 1$. Note that $0 \le d_i \le 1$.

Finally, we calculate the grounding score $g_i$ by applying an exponential decay function to $d_i$: $g_i = e^{\frac{-d_i}{1-d_i}}$. Note that the score rewards cases where $\mathbf{b}_{t_i}$ and $\mathbf{b}_{a_i}$ are close, or horizontally/vertically aligned (due to lower Manhattan Distance) with the reward dropping exponentially with distance. The exponential decay function was demonstrated to best represent positional information in unimodal text in \citet{kerple2022}, and extended to multimodal documents in \citet{docgraphlm2023}.

\subsection{Type-aware surface similarity}
To measure $m_i$, the surface match score between $t_i$ and $a_i$, we follow the below criteria:

\begin{enumerate}
\item If $t_i$ is textual\footnote{See Appendix \ref{app:types} for additional details.}, we use the NLS metric. 
\item If $t_i$ is numeric, we use a binary score that indicates whether the predicted answer matches the ground truth exactly. We allow some flexibility in the match, for example numbers scaled by 100, thousand, million, or billion are considered a match. This is to account for different expressions of percentages, basis points, financial metrics, etc. 
\item If $t_i$ is composed of both textual and numeric characters, we first create substrings $\text{num}_{a_i}$, $\text{str}_{a_i}$, $\text{num}_{t_i}$, and $\text{str}_{t_i}$ by extracting the numeric and non-numeric characters of $a_i$ and $t_i$, respectively. Next, we calculate the number-based and text-based scores for each substring according to the above criteria. The final score is a weighted harmonic mean of the two subscores: $\frac{11}{\frac{10}{\text{num\_score}_i}+\frac{1}{\text{str\_score}_i}}$.\footnote{See Appendix \ref{app:tuning} for additional details.} Note that the model has to get the numeric part of the answer correctly to be rewarded higher.
\end{enumerate}

\subsection{Composite metric}
\label{sec:composite}
Given the mutimodal grounding score $g_i$ and type-aware match score $m_i$, we propose the following composite score parameterized by $\alpha$:

\begin{equation}
s_i = \alpha m_i + (1-\alpha) g_i
\end{equation}

Note that $\alpha = 0$ yields the grounding score and $\alpha = 1$ yields the type-aware match score. The configurability of the $\alpha$ parameter allows users to tune it on a validation set of their choice, or, as we will show in Section \ref{sec:calibration}, to optimize it such that it rewards well-calibrated outputs. 

\section{Experiments}

\label{sec:data}
Given the composite score proposed in Section \ref{sec:composite}, we investigate the impact of groundedness on four prominent Document VQA benchmarks.

\textbf{DocVQA} \cite{docvqa2021} is a visual question-answering (VQA) dataset designed specifically for document images. It contains over 12,000 document images sourced from scanned business forms, reports, and invoices, among others. The dataset is structured with over 50,000 question-answer pairs, and questions are broken down into 9 categories, indicating the context of the correct answer (e.g. ``Free\_text'', ``Layout'', ``Figure/Diagram'', etc.). This breakdown is not available for the test collection. Therefore we determine the type of each question using GPT-4o \cite{gpt4o}\footnote{Please see Appendix \ref{app:qtypes} for details.}. Next we remove questions in the ``Yes/No'' category to filter potentially abstractive questions. This results in 5,130 questions in the final dataset. 

\textbf{InfographicVQA}. \cite{infographicvqa2022} is a dataset aimed at visual question answering over complex infographic documents. The dataset includes over 5,000 infographic images and over 30,000 questions that require reasoning over text, charts, and images embedded within the infographic. We filter multi-piece answers from the test collection, resulting in 3,272 samples. 

\textbf{MP-DocVQA} \cite{mpdocvqa2023} focuses on multi-page documents. It consists of over 46,000 question-answer pairs from 6,000 multi-page documents. We use 5,019 questions in the test set.

\textbf{DUDE} \cite{dude2023} is a document understanding dataset focused on structured documents such as forms, invoices, and tables. It includes around 5,000 documents and 41,000 question-answer pairs. We limit the test collection to single-piece extractive questions, resulting in 2,552 samples.

For each sample in each dataset, we calculate the NLS as well as the composite score, with $\alpha$ set to increments of $0.05$ in the $[0, 1]$ range.

\section{Analysis}
Throughout most of our experiments, we set $\alpha = 0.25$, as it proves optimal based on the calibration analysis provided in Section \ref{sec:calibration}. Since $\alpha$ is optimized on the DUDE dataset, we have not included this dataset in any of the analyses that use this optimal value for $\alpha$.

\subsection{Leaderboard analysis}
We first analyze how \metricName can affect the rankings produced by Document VQA benchmarks. Figure \ref{fig:docvqa_reranking} illustrates this using the top 10 models\footnote{As of September 2024.} on the DocVQA leaderboard. The leftmost column of the figure shows the original ANLS-based ranking\footnote{Note that our ANLS-based rankings could be slightly different from the leaderboard, since we have filtered the questions per Section \ref{sec:data}.}. The second column shows how the ranking changes if we switch to \metricName with $\alpha = 0.25$. As the figure shows, human performance and QWen2-VL \cite{qwen2vl2024} remain stable, but all other models move by at least one position on the leaderboard. The middle segment of the figure shows how the models would rank based on the type of question. Certain question types such as ``Figure/Diagram'' and ``Table/List'' offer little volatility, but for questions that fall under ``Handwritten'' or ``Other'', the volatility is higher.\footnote{An example of a question classified as ``Other'' is: ``What does GCC stand for?'' requiring the model to infer that an acronym mentioned on one part of a page is related to an entity mentioned on a different part. This category of questions constitutes about 0.2\% of the DocVQA dataset, and can be considered negligible.} The middle segment of the figure also shows that some models such as SMoLA-PaLI-X \cite{smolapalix2024} are better at answering questions based on ``Free\_text'' contexts, whereas they struggle with ``Table/List'' questions compared to other models\footnote{Note that all changes in rankings have been visualized with regard to the universal ranking of the leaderboard (left-most column), as opposed to the original rank within that question type. This is because, as of September 2024, the rank of the top 5 models on the DocVQA leaderboard was largely consistent with their rank within each question type. Given the relative stability of the rankings and to simplify the graphic, we chose to visualize all movements against the universal ranking.}.

The right segment of the figure shows the rerankings broken down by the type of answer. As expected, textual answers offer the closest ranking to the original one produced by ANLS, whereas numeric and hybrid answers perturb the ranking of the leaderboard. Notably, humans remain the top performer for textual and hybrid answers, but fall behind two other models in the numeric category. This can be attributed to the human tendency to rephrase certain entities such as numbers and dates. For example, in Question \#3027, the ground truth answer ``(16.1\%)'' is rephrased as ``-16.1\%'' by human respondents, and for question \#3290, ``1,700'' is modified as ``about 1,700''. 

\input{latex/figdocvqarank}
\input{latex/figallrank}

Figure \ref{fig:atypes} shows the correlation between rankings produced by ANLS and by our composite score with $\alpha=0.25$. Following \citet{benchmarks2024}, we calculate the correlation based on a two-tailed Kendall's $\tau$ analysis. Note that the y-axis on Figure \ref{fig:atypes} begins at $0.70$. As the figure shows, questions with textual answers are the least affected by switching to our score, but numeric and hybrid answers impact the ranking by a larger margin. This is expected as the text-only version of our score is the closest to ANLS. Of the three benchmarks shown in the figure, InfographicVQA is most affected by our score, whereas DocVQA and MP-DocVQA retain a strong correlation with their original rankings. As evidenced by Figure \ref{fig:docvqa_reranking}, this strong correlation does not indicate a stable leaderboard, but one where the models move by $\pm d$, where $d$ is a small number.

\subsection{Question type analysis}
Figure \ref{fig:qtypes} shows the correlation between our composite score and the original ranking of the DocVQA leaderboard for each question type. As expected, moving from small values of $\alpha$ (weighing groundedness more that type-aware similarity) to large values (weighing type-aware similarity more than groundedness), moves the rankings closer to the original ANLS ranking. This is especially true of the ``Free\_text'' category, where our score comes closest to ANLS. Once again, ``Other'' is the outlier category, which can be safely ignored due to its small sample size. The remaining categories show a similar trend, further establishing that groundedness is not accounted for in ANLS-based rankings. 

\input{latex/figdocvqaqtypes}



\subsection{Association with calibration}
\label{sec:calibration}
The DUDE dataset provides the confidence scores produced by each model (when available). This enables the benchmark to report Expected Calibration Errors (ECE) \cite{ece2015}, indicating if the models are wrongfully over or under-confident about the accuracy of their output. We use this metric to determine whether our proposed score can account for accuracy through calibratedness. To do this, we map the score at various $\alpha$'s against the calibration error of each model, and calculate the Pearson-R correlation between the two. The results are displayed in Figure \ref{fig:calib}. As the figure shows, at small values of $\alpha$ (focusing on groundedness), there is a negative correlation with ECE, indicating that a higher score is correlated with a lower ECE. As $\alpha$ increases and the score shifts towards surface similarity, the association moves towards positive, crossing $0$ around $\alpha = 0.5$. This trend can be observed for all categories of questions except ``Textual'' questions, which enforce surface similarity at all $\alpha$ values. The optimal value for $\alpha$, which minimizes the correlation with ECE across most categories lies at around $\alpha=0.25$. 

\input{latex/figcalib}

\subsection{Association with robustness}
Next, we inspect the association between \metricName and the robustness of a given model. 
Robustness is not a formally defined term in the Document VQA field, but can be interpreted as a model's consistent performance across different settings, benchmarks, and sample types. Therefore, we define robustness as the volatility\footnote{See Appendix \ref{app:vol} for additional details.} of a model's ranking when evaluated on various subsets of questions (e.g. textual, numeric, hybrid, or all questions at once). We plot this volatility against the volatility of a model's scores, using the DocVQA, MP-DocVQA, and InfographicVQA benchmarks. Figure \ref{fig:robust} shows the results using ANLS as well as \metricName with $\alpha=0.25$. Each dot represents one model, with red dots representing models evaluated using ANLS, and blue dots representing models evaluated by \metricName. As the regression lines in the figure show, both approaches maintain a positive trend between the volatility in scores and rankings. In other words, models with stable rankings tend to have stable scores as well. However, the positive trend is stronger for our score compared to ANLS, with a small but statistically significant regression coefficient of 0.58 (compared to ANLS's 0.33).  

\input{latex/figrobust}

Next, to present a qualitative view of how our score can reward robust models, we calculate a robustness score for each model in the DocVQA benchmark. To do this, we scale a model's rank volatility by its median rank. This ensures that if a model is stable across rankings, it receives a high robustness score, unless it is a generally poor performing model (e.g. a model that comes last in all rankings). Table \ref{tab:robust} lists the top-5 models identified using this technique. The ANLS-based models reflect the default ranking of the DocVQA leaderboard, with Humans leading the group, followed by Large MLMs such as QWen2-VL \cite{qwen2vl2024} and InternVL2-Pro/InternVL-1.5 \cite{internvl2024}. 

In contrast, our score produces a ranking that includes a Small MLM, namely, Arctic-TILT \cite{tilt2024}. As of October 2024, this model is ranked 11 on the DocVQA leaderboard, above all other Small MLMs and a few Large MLMs. In addition, it is ranked 1st on the MP-DocVQA and DUDE leaderboards. No other models listed in the ANLS column show the same level of cross-benchmark robustness. Similarly, Molmo-72B \cite{molmo2024} is 4th on the InfographicVQA benchmark. The strong cross-benchmark rankings indicate that our method can generate rankings that reward robust models. 

\input{latex/tabqual}
\input{latex/tabrobust}

\subsection{Human evaluation}
We used human judgment to assess the validity of our scores compared to ANLS. To do this, we used data from three benchmarks: DocVQA, MP-DocVQA, and InfographicVQA. In each benchmark, we sampled questions and a pair of answers produced by two models, indicated by model A and model B (different models could be 
selected for each sample). We limited the samples to cases where model A's NLS score was higher than B, but \metricName scored B higher than A, or vice versa. We sampled up to 100 such question-answers triplets from each benchmark\footnote{Some datasets had fewer qualifying triplets.}. Three researchers were presented with these triplets, as well as the ground truth answer, and asked which model they thought should be scored higher. The annotations produced a mean Cohen's $\kappa$ of $0.82$, indicating a high level of agreement. We filtered the annotations to those on which at least two annotators agreed. This resulted in 28 samples for DocVQA, 86 samples for MP-DocVQA, and 66 samples for InfographicVQA.

Figure \ref{fig:userstudy} shows the annotators' agreement rates with NLS versus our score. The ``Neither'' bucket indicates that the annotators believed the models should have been scored equally. As the figure shows, annotators agreed with \metricName in the majority of cases across all three benchmarks, indicating that our approach is better aligned with human judgment. We observe that InfographicVQA, which yielded the highest rate of agreement with NLS, contains the largest number of misspelled numbers. This could be a result of the complex layout and design of infographics. 

\input{latex/figuserstudy}

\section{Conclusion}
In this study, we showed how popular evaluation metrics such as ANLS can miss important nuances when used to analyze Document VQA models. Instead, we proposed \metricName, a new metric that is sensitive to the groundedness of the models' outputs. Through extensive analyses, we showed how \metricName is better aligned with human judgement as well as the calibratedness of the models. Our analyses also showed that rankings produced by \metricName were better indicators of a model's robustness across question types and in different benchmarks. Our studies demonstrate the importance of groundedness in the performance and assessment of Document VQA models. We hope that in addition to presenting a new evaluation method, our study inspires researchers to develop better grounded Document VQA models.

\section{Limitations}
The analyses performed in this paper were all conducted on single-span, extractive answers. To extend the grounding mechanism to multi-span answers, the matching algorithm would need to handle an arbitrary number of partitions, unless the benchmark specifically identifies each span in its test set annotations. Determining groundedness on abstractive questions is a challenging task that is outside of the scope of this study. 

The methodology proposed in this study does not account for semantic categories that go beyond textual/numeric/hybrid forms, such as currencies, dates, timestamps, etc. each of which come with nuances that can be mishandled by solely considering surface similarity. 

Since the $\alpha$ parameter was tuned on the DUDE dataset, it was excluded from some of the other analyses. The remaining benchmarks (DocVQA, MP-DocVQA, and InfographicVQA) are all released as part of the same suite of tasks, with the first two datasets being based on the same collection of documents. This can lead to biases in the analyses that are based solely on these three benchmarks. However, none of these benchmarks provided access to the confidence scores produced by the models, and therefore could not be used to tune $\alpha$.

Lastly, the grounding algorithm mentioned in Section \ref{sec:grounding} relies on the accuracy of the reading order of each page, as presented in the OCR output. As \citet{order2023} point out, this can often be noisy or misleading. Appendix \ref{app:grounding} offers an alternative, more generalizable, yet slower solution using a walk over the $\beta$-skeleton presentation of each page.

\section{Acknowledgments}
Armineh Nourbakhsh, Pranav Shetty, and Sameena Shah's work is supported by JPMorgan Chase \& Co. This paper was prepared for informational purposes by the Artificial Intelligence Research group of JPMorgan Chase \& Co and its affiliates (``JP Morgan''), and is not a product of the Research Department of JP Morgan. JP Morgan makes no representation and warranty whatsoever and disclaims all liability, for the completeness, accuracy or reliability of the information contained herein. This document is not intended as investment research or investment advice, or a recommendation, offer or solicitation for the purchase or sale of any security, financial instrument, financial product or service, or to be used in any way for evaluating the merits of participating in any transaction, and shall not constitute a solicitation under any jurisdiction or to any person, if such solicitation under such jurisdiction or to such person would be unlawful.

\bibliography{acl_latex}

\appendix

\input{latex/appgrounding}

\input{latex/appbboxes}

\input{latex/appqtypes}

\input{latex/appreranking}

\input{latex/appdocvqaqtypes}

\input{latex/appdocvqaatypes}

\input{latex/appmpdocvqaatypes}

\input{latex/appinfoatypes}

\input{latex/appdudeatypes}

\input{latex/appatypes}
\end{document}

%% file: latex/figexample.tex
\begin{figure*}[t!]
    \centering
    \begin{subfigure}[t]{0.5\textwidth}
        \centering
        \includegraphics[height=1.2in]{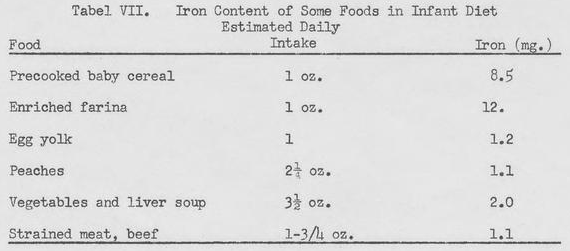}
        \caption{Tabular snippet.}
        \label{fig:snippetstab}
    \end{subfigure}%
    ~ 
    \begin{subfigure}[t]{0.5\textwidth}
        \centering
        \includegraphics[height=1.2in]{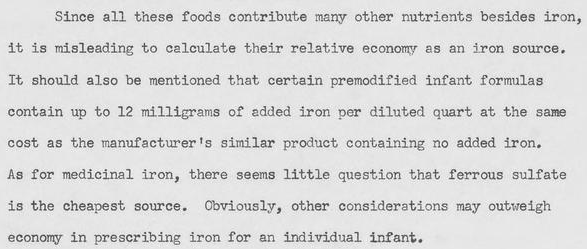}
        \caption{Text snippet.}
        \label{fig:snippetspar}
    \end{subfigure}
    \caption{Two excerpts from an image document from the DocVQA dataset \cite{docvqa2021}.}
    \label{fig:snippets}
\end{figure*}

%% file: latex/tabexample.tex
\begin{table*}[t!]
\caption{Two example questions based on the snippets in Figure \ref{fig:snippets}. The ``NLS'' column shows the score awarded to hypothetical answers for each question using the NLS metric \cite{docvqa2021}. In ``Ours'', we show how our proposed score is calculated.}
\centering
\resizebox{0.95\textwidth}{!}{%
\begin{tabular}{l|l|l|c|c|ccccccc}
\hline
\multicolumn{1}{c|}{\multirow{3}{*}{\textbf{Question}}}                                                                  & \multirow{3}{*}{\textbf{Context}} & \multicolumn{1}{c|}{\multirow{3}{*}{\textbf{\begin{tabular}[c]{@{}c@{}}GT\\ Answer\end{tabular}}}} & \multirow{3}{*}{\textbf{\begin{tabular}[c]{@{}c@{}}Predicted\\ Answer\end{tabular}}} & \multirow{3}{*}{\textbf{NLS}} & \multicolumn{7}{c}{\textbf{\metricName (Ours)}}                                                                                                                                                                                                                                                                                                                                                                                                                                                                                                            \\ \cline{6-12} 
\multicolumn{1}{c|}{}                                                                                                    &                                   & \multicolumn{1}{c|}{}                                                                              &                                                                                      &                               & \multicolumn{3}{c|}{\textbf{\begin{tabular}[c]{@{}c@{}}Match\\ Score\end{tabular}}}                                                                                                                         & \multicolumn{3}{c|}{\textbf{\begin{tabular}[c]{@{}c@{}}Grounding\\ Score\end{tabular}}}                                                                                                                                      & \multirow{2}{*}{\textbf{\begin{tabular}[c]{@{}c@{}}Composite\\Score\\ ($\alpha = 0.25$)\end{tabular}}} \\ \cline{6-11}
\multicolumn{1}{c|}{}                                                                                                    &                                   & \multicolumn{1}{c|}{}                                                                              &                                                                                      &                               & \multicolumn{1}{c|}{\textbf{\begin{tabular}[c]{@{}c@{}}Text\\ Score\end{tabular}}} & \multicolumn{1}{c|}{\textbf{\begin{tabular}[c]{@{}c@{}}Num\\ Score\end{tabular}}} & \multicolumn{1}{c|}{\textbf{Agg.}} & \multicolumn{1}{c|}{\textbf{\begin{tabular}[c]{@{}c@{}}Horizontal\\Distance\end{tabular}}} & \multicolumn{1}{c|}{\textbf{\begin{tabular}[c]{@{}c@{}}Vertical\\Distance\end{tabular}}} & \multicolumn{1}{c|}{\textbf{Agg.}} &                                                                                                 \\ \hline
\multirow{2}{*}{\begin{tabular}[c]{@{}l@{}}How many mgs of\\ iron is in enriched farina?\end{tabular}}                   & \multirow{2}{*}{Figure \ref{fig:snippetstab}}              & \multirow{2}{*}{12}                                                                                & 26                                                                                   & 0.5                           & \multicolumn{1}{c|}{-}                                                             & \multicolumn{1}{c|}{0.0}                                                          & \multicolumn{1}{c|}{0.0}           & \multicolumn{1}{c|}{-}                                                                    & \multicolumn{1}{c|}{-}                                                                      & \multicolumn{1}{c|}{0.0}           & 0.0                                                                                             \\ \cline{4-12} 
 &                                   &                                                                                                    & 8.5                                                                                  & 0.0                           & \multicolumn{1}{c|}{-}                                                             & \multicolumn{1}{c|}{0.0}                                                          & \multicolumn{1}{c|}{0.0}           & \multicolumn{1}{c|}{$\sim{}$0.0}                                                                  & \multicolumn{1}{c|}{0.2}                                                                    & \multicolumn{1}{c|}{0.02}           & 0.01                                                                                            \\ \hline
\multirow{2}{*}{\begin{tabular}[c]{@{}l@{}}How much added iron\\ do premodified infant\\ formulas contain?\end{tabular}} & \multirow{2}{*}{Figure \ref{fig:snippetspar}}              & \multirow{2}{*}{\begin{tabular}[c]{@{}l@{}}up to 12\\ milligrams\end{tabular}}                     & \multicolumn{1}{l|}{\begin{tabular}[c]{@{}l@{}}up to\\ 12 mgs\end{tabular}}          & \multicolumn{1}{l|}{0.58}     & \multicolumn{1}{c|}{0.59}                                                          & \multicolumn{1}{c|}{1.0}                                                          & \multicolumn{1}{c|}{0.74}          & \multicolumn{1}{c|}{0.0}                                                                  & \multicolumn{1}{c|}{0.0}                                                                    & \multicolumn{1}{c|}{1.0}           & 0.93                                                                                            \\ \cline{4-12} 
&                                   &                                                                                                    & \multicolumn{1}{l|}{\begin{tabular}[c]{@{}l@{}}up to 1z\\ milligrams\end{tabular}}   & \multicolumn{1}{l|}{0.95}     & \multicolumn{1}{c|}{0.94}                                                          & \multicolumn{1}{c|}{0.0}                                                          & \multicolumn{1}{c|}{0.0}           & \multicolumn{1}{c|}{0.0}                                                                  & \multicolumn{1}{c|}{0.0}                                                                    & \multicolumn{1}{c|}{1.0}           & 0.75                                                                                             \\ \hline
\end{tabular}
}
\label{tab:example}
\end{table*}

%% file: latex/figdocvqarank.tex
\begin{figure*}[h]
    \centering
    \includegraphics[width=0.8\textwidth]{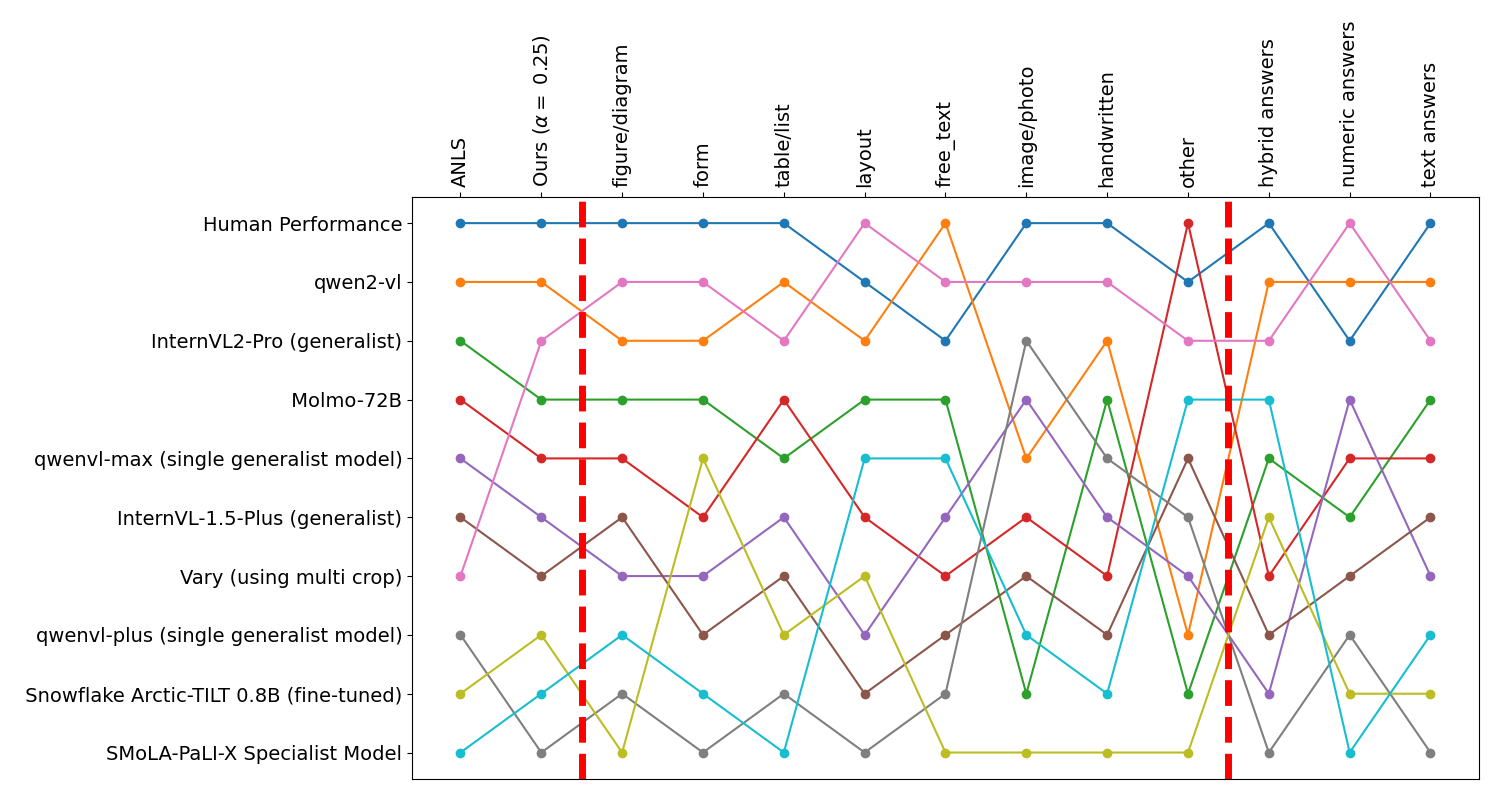}
    \caption{The rankings of the top 10 models on the DocVQA leaderboard, before and after applying our composite score with $\alpha = 0.25$. Left segment: Rankings based on ANLS versus our score. Middle segment: Our rankings broken down by question type. Right segment: Our rankings broken down by answer type.}
    \label{fig:docvqa_reranking}
\end{figure*}

%% file: latex/figallrank.tex
\begin{figure}[h]
    \centering
    \includegraphics[width=\columnwidth]{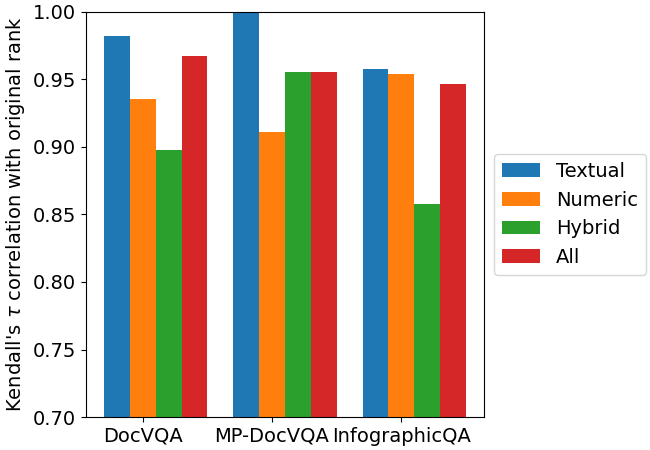}
    \caption{The correlation between the rankings produced by our method (with $\alpha=0.25$) and the original ANLS-based ranking, broken down by the type of answer. All $\tau$ values are significant at $p \ll 0.05$.}
    \label{fig:atypes}
\end{figure}

%% file: latex/figdocvqaqtypes.tex
\begin{figure}[h]
    \centering
    \includegraphics[width=\columnwidth]{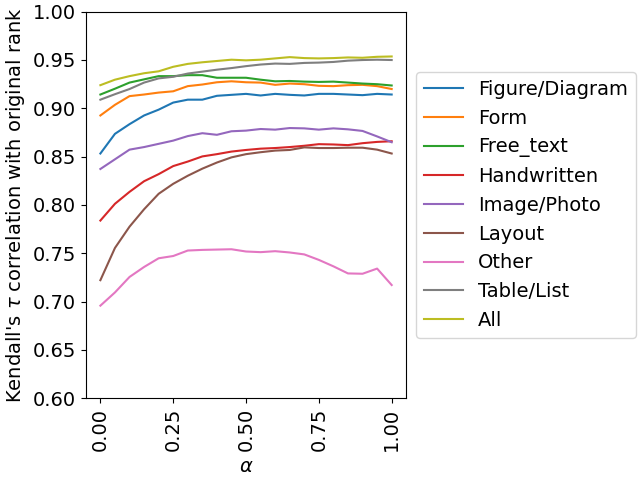}
    \caption{Kendall's $\tau$ rank correlation with the original DocVQA leaderboard, broken down by question types. All $\tau$ values are significant at $p \ll 0.05$.}
    \label{fig:qtypes}
\end{figure} 

%% file: latex/figcalib.tex
\begin{figure}[h]
    \centering
    \includegraphics[width=0.8\columnwidth]{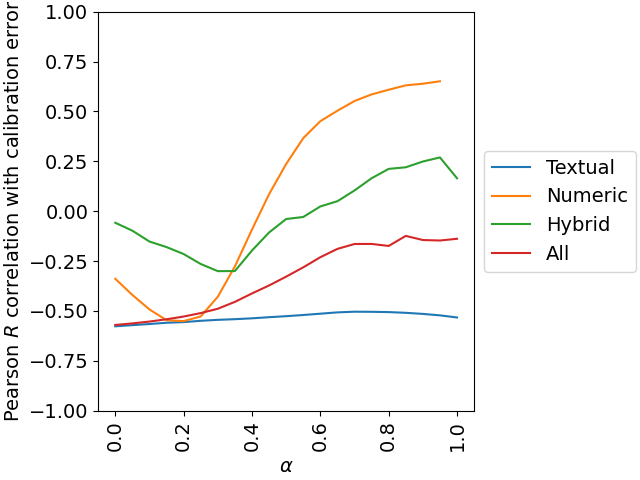}
    \caption{Pearson $R$ correlation with the calibration error of models based on the DUDE leaderboard, broken down by answer type.}
    \label{fig:calib}
\end{figure}

%% file: latex/figrobust.tex
\begin{figure}[h]
    \centering
    \includegraphics[width=0.8\columnwidth]{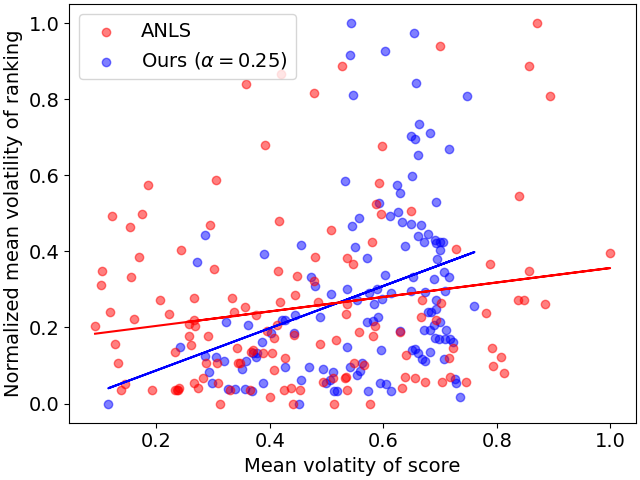}
    \caption{The mean volatility of each model's score versus its ranking. Red dots represent ANLS scores and blue dots represent \metricName with $\alpha = 0.25$.}
    \label{fig:robust}
\end{figure}

%% file: latex/tabqual.tex
\begin{table*}[ht]
\caption{Five samples from the human preference study, showing cases where the human judges preferred our score, NLS, or neither scores. In the latter case, the human judges preferred equal scores for Models A and B.}
\resizebox{\textwidth}{!}{%
\begin{tabular}{l|l|l|l|l|l}
\hline
\textbf{Dataset} & \textbf{Question} & \textbf{GT} & \textbf{Model A} & \textbf{Model B} & \textbf{\begin{tabular}[c]{@{}l@{}}Human pick\end{tabular}} \\ \hline
\multirow{3}{*}{DocVQA}    & \multirow{3}{*}{\begin{tabular}[c]{@{}l@{}}What is the vitamin A requirement\\(in I.U.) for a 'lactating' mother ?\end{tabular}} & \multirow{3}{*}{\begin{tabular}[c]{@{}l@{}}``1,000 i.u. plus \\ basic requirements''\end{tabular}}   & ``basic requirements''  & ``1,000'' &  \multirow{3}{*}{\metricName}  \\ \cline{4-5}
&  &  & NLS: 0.54  & NLS: 0.0 &      \\ \cline{4-5}
&   &  & Ours: 0.0  & Ours: 0.62 &   \\ \hline
\multirow{3}{*}{MP-DocVQA} & \multirow{3}{*}{What is the day and date of Meeting?}    & \multirow{3}{*}{\begin{tabular}[c]{@{}l@{}}``thursday 22\\ october''\end{tabular}} & ``thursday'' & \begin{tabular}[c]{@{}l@{}}``saturday 24 october''\end{tabular}  &  \multirow{3}{*}{\metricName}   \\ \cline{4-5}
& &  & NLS: 0.0 & NLS: 0.74 & \\ \cline{4-5}
 &  & & Ours: 0.81  & Ours: 0.25  &  \\ \hline
 \multirow{3}{*}{InfographicVQA} & \multirow{3}{*}{{\begin{tabular}[c]{@{}l@{}}Which age group uses social media\\the most?\end{tabular}}}    & \multirow{3}{*}{\begin{tabular}[c]{@{}l@{}}``18-29 \\year olds''\end{tabular}}
 & ``18-29 group'' & \begin{tabular}[c]{@{}l@{}}``18-24 year olds''\end{tabular}  & \multirow{3}{*}{\metricName}    \\ \cline{4-5}
& &  & NLS: 0.53 & NLS: 0.93 & \\ \cline{4-5}
 &  & & Ours: 0.98  & Ours: 0.0  &  \\ \hline

  \multirow{3}{*}{DocVQA} & \multirow{3}{*}{{\begin{tabular}[c]{@{}l@{}}What is the date of the letter?\end{tabular}}}    & \multirow{3}{*}{\begin{tabular}[c]{@{}l@{}}``august 1, 1983''\end{tabular}}
  & \begin{tabular}[c]{@{}l@{}}``The date of the letter \\is August 1, 1983.''\end{tabular} & \begin{tabular}[c]{@{}l@{}}``August 1983''\end{tabular}  &  \multirow{3}{*}{Neither}   \\ \cline{4-5}
& &  & NLS: 0.0 & NLS: 0.78 & \\ \cline{4-5}
 &  & & Ours: 0.97  & Ours: 0.0  &  \\ \hline

  \multirow{3}{*}{InfographicVQA} & \multirow{3}{*}{{\begin{tabular}[c]{@{}l@{}}What is the estimated number\\(in billions) of social media\\users around the globe by 2019?\end{tabular}}}    & \multirow{3}{*}{\begin{tabular}[c]{@{}l@{}}``2.72''\end{tabular}}                         & 
 ``\#infographic'' & \begin{tabular}[c]{@{}l@{}}``2. 72''\end{tabular}  &   \multirow{3}{*}{ANLS}  \\ \cline{4-5}
& &  & NLS: 0.0 & NLS: 0.8 & \\ \cline{4-5}
 &  & & Ours: 0.0  & Ours: 0.0  &  \\ \hline
\end{tabular}
}
\label{tab:qual}
\end{table*}

%% file: latex/tabrobust.tex
\begin{table}[]
\centering
\caption{Top-5 models based on robustness rankings produced by ANLS versus our score (with $\alpha = 0.25$).}
\resizebox{0.8\columnwidth}{!}{%
\begin{tabular}{ll|ll}
\hline
\multicolumn{2}{c|}{\textbf{ANLS}} & \multicolumn{2}{c}{\textbf{\metricName}} \\ \hline
1          & Human                 & 1     & Human                     \\ \hline
2          & QWen2-VL              & 2     & QWen2-VL                  \\ \hline
3          & InternVL2-Pro         & 3     & InternVL2-Pro               \\ \hline
4          & QWenVL-Max            & 4     & Molmo-72B                 \\ \hline
5          & InternVL-1.5-Plus              & 5     & Snowflake Arctic-TILT     \\ \hline
\end{tabular}
}
\label{tab:robust}
\end{table}

%% file: latex/figuserstudy.tex
\begin{figure}[h]
    \centering
    \includegraphics[width=0.85\columnwidth]{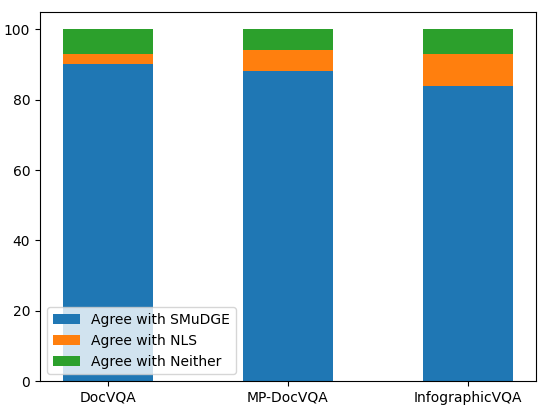}
    \caption{Human preference for pairwise rankings produced by NLS versus \metricName (with $\alpha = 0.25$).}
    \label{fig:userstudy}
\end{figure}

%% file: latex/appgrounding.tex
\section{A $\beta$-skeleton based grounding algorithm}
\label{app:grounding}
Algorithm \ref{alg:beta} describes a possible way to ground a model output $O$ within a page, without apriori access to the reading order. First, a page is represented by a $\beta$-skeleton graph, similar to \citet{rope2021}. Next, the first and last tokens of $O$ are matched to the page by finding all nodes (i.e. tokens) on the graph that have a Levenshtein similarity to the first and last token, beyond a threshold $T$. Lastly, all possible paths between such nodes are found, and the path with the highest NLS to $O$ is selected as the matching path.

A threshold can be set on the score of the best matching path $S$, below which the path is considered a mismatch and therefore no effective matches are found on the page, e.g. in cases when the model has hallucinated the output.

This algorithm ensures that any path that is matched to $O$ is within a contiguous 2-D walk on the page, without the need for information related to reading order. A major downside of this algorithm is its quartic time complexity, which can be improved by caching partial paths. Nevertheless, we decided to use a simpler algorithm that relies on the reading order provided by OCR tools. 

\begin{algorithm}[h!]
\caption{$\beta$-skeleton walk for placing a sequence of tokens within a page.}
\label{alg:beta}
  \begin{algorithmic}[1]
  \Statex{\textit{// $\beta$-skeleton representation of a page}}
  \STATE $\text{Input: } G = (N, V)$
  \Statex{\textit{// Matching target: a sequence of tokens}}
  \STATE $\text{Input: }O = o_1, o_2, \cdots, o_n$
  \Statex{\textit{// Threshold for token similarity}}
  \STATE $\text{Input: }T$
  \Statex{\textit{// Best path on the graph that matches $O$}}
  \STATE $\text{Output: }P$
  \Statex{\textit{// The similarity of the best path to $O$}}
  \STATE $\text{Output: }S$
  \Statex{\textit{// Create empty indices of all possible paths over the graph, starting from $o_1$ ending in $o_n$. }}
  \STATE $p_s \gets \{\}$
  \STATE $p_e \gets \{\}$
  \FOR{$i \in \{1,\dots, |N|\}$}
    \STATE $s_{i1} = \text{NLS}(N_i, o_1)$
    \STATE $s_{in} = \text{NLS}(N_i, o_n)$
    \IF{$s_{i1} > T$}
        \STATE $\text{append}(p_s, n_i)$
    \ENDIF
    \IF{$s_{in} > T$}
        \STATE $\text{append}(p_e, n_i)$
    \ENDIF 
    \Statex{\textit{// Search all possible paths and select the one with the highest score}}
    \FOR{$p_j \in p_s$}
        \FOR{$p_k \in p_e$}
            \FOR{$\text{path} \in \text{paths}(p_j \rightarrow p_k)$}
                \IF{$\text{NLS}(\text{path}, O) > S$}
                    \STATE $S \gets \text{NLS}(\text{path}, O)$
                    \STATE $P \gets \text{path}$
                \ENDIF
            \ENDFOR
        \ENDFOR
    \ENDFOR
  \ENDFOR
  \end{algorithmic}
\end{algorithm}

%% file: latex/appbboxes.tex
\section{Additional experimental details}
\subsection{Merging bounding boxes}
\label{app:bboxes}
A sequence of bounding boxes can be merged by finding the left-most, top-most, right-most, and bottom-most corners in the sequence in order to create a new bounding box. If all bounding boxes in the sequence form a contiguous segment, merging them would yield their union. However, if the bounding boxes are in disparate locations, this simple merging algorithm will not yield their union, and will cover additional areas. As an example, if a ground truth answer spans two lines, covering the second half of one line and the first half of the next, the merging algorithm will create a bounding box that covers both lines in full. Despite this limitation, we use this algorithm because we are only interested in measuring the distance between the resulting bounding boxes based on their centroids. 

\subsection{Normalizing the distance}
\label{app:norm}
Given the ground truth bounding box $\mathbf{b}_{t_i}$ and the predicted bounding box $\mathbf{b}_{a_i}$, our goal is to measure the distance between the centroids of the bounding boxes. In our proposed formulation, this distance is normalized by the width and height of the page, namely $\texttt{width}_i$ and $\texttt{height}_i$. This is not the only possible option for normalizing the distance. For example, the distance can be normalized by the width/height of the ground truth bounding box $\mathbf{b}_{t_i}$, or the average size of the ground truth and predicted bounding boxes $\mathbf{b}_{t_i}$ and $\mathbf{b}_{a_i}$. 

Each option offers advantages and disadvantages, which we will demonstrate using examples. For simplicity, we will suppose that the height of all bounding boxes is similar, and focus on width only. 

Normalizing the bounding boxes by the width of $\mathbf{b}_{t_i}$ over-penalizes models that provide short answers compared to the ground truth, and under-penalizes models that provide longer answers compared to ground truth. 
A real example from the DocVQA dataset is the question ``What decides the selection of terms of Committee members?'' The ground truth answer is ``decided by a lottery'', whereas some models may produce ``lottery'' and some may produced ``will be decided by a lottery''. We would want the grounding distance to be consistently low for these variations. But once normalized by the width of ground truth, the first model will be over-penalized and the second model will be under-penalized. 

An alternative is to normalize the widths by the average widths of $\mathbf{b}_{t_i}$ and $\mathbf{b}_{a_i}$. While this formulation does not suffer from sensitivity to the variety of sizes, it does ignore the sizes of the bounding boxes relative to the size of the page. For example, consider a page with width $1000$. On this page, these two scenarios produce the same distance of $1$: 

Scenario A: $\mathbf{b}_{t_i}$ spans $[0-500]$ and $\mathbf{b}_{a_i}$ spans $[500-1000]$. The average width is $500$. The centroids are at $250$ and $750$, respectively. Therefore the centroids are at a raw distance of $500$ and a normalized distance of $1.$

Scenario B: $\mathbf{b}_{t_i}$ spans $[0-50]$ and $\mathbf{b}_{a_i}$ spans $[50-100]$. The average width is $50$. The centroids are at $25$ and $75$, respectively. Therefore the centroids are at a raw distance of $50$ and a normalized distance of $1.$

On the one hand, it can be argued that it is fair for the distances to be equal, as the bounding boxes are adjacent in both scenarios. On the other hand, it can be argued that the distance should scale with the width of the bounding boxes compared to the width of the page, because locating a small bounding box on a large page is more difficult than a large bounding box on a small page.

In contrast, given that the width of the page is $1000$, our proposed method would produce a distance of $0.5$ for Scenario A and $0.05$ for Scenario B. This is an interpretable metric, as it indicates that the two bounding boxes are within $50\%$ of the width of the page in Scenario A, and $5\%$ in Scenario B. It can of course be argued that our formulation is too sensitive to the scale of the page. Our proposed score is indeed not perfect, but we consider it to be an interpretable ``layout-agnostic'' metric that can be easily calculated across all samples.

We thank our reviewer for suggesting these alternative options. 

\subsection{Determining the semantic type of the predicted answer}
\label{app:types}

To classify a string of characters $s$ as numeric, textual, or hybrid, we follow the below algorithm:

\begin{enumerate}
\item If every character in $s$ is a digit, then $s$ is numeric.
\item If every character in $s$ is alphabetical, then $s$ is textual.
\item Otherwise $s$ is hybrid.
\end{enumerate}

Note that this simple algorithm renders a large portion of strings such as ``1,700'' or ``(8)'' as hybrid. This is not detrimental to \metricName, as it still favors the accuracy of numbers against non-numeric characters by a factor of 10 to 1 (see Appendix \ref{app:tuning}). 

Note that hybrid strings are split into numeric sequences and non-numeric sequences, e.g. ``1,700'' is split into ``1700'' and ``,'' and each part is evaluated separately before being combined in the weighted harmonic mean.

\subsection{Tuning the weights for the numeric score and the text score}
\label{app:tuning}

Setting the weight of $\text{num\_score}_i$ to 1 would mean that the numeric and text components of an answer would have equal importance, which is indeed not valid. For example if the ground truth is ``12 milligrams'', then the answers ``2 milligrams'' and ``12 milligram'' should not receive equal scores, as the former is quantitatively incorrect, but the latter has a simple typo. On the other hand, setting a very high weight for $\text{num\_score}_i$ can be problematic. For example if the ground truth is ``12 mgs'' and the predicted answer is ``12 ms'', we would need to properly penalize the text component, because ``ms'' stands for ``milliseconds'' and not ``milligrams''. 

Therefore we tuned the weight of $\text{num\_score}_i$ against $\text{str\_score}_i$ by testing values in the set $\{1, 10, 100, 1000\}$. The tuning was performed on a subsample of 100 hybrid answers from the DocVQA validation set, and validated by three human annotators. Each annotator was presented with answer/ground-truth pairs and the four variations of the score calculated using the four values in $\{1, 10, 100, 1000\}$. The annotators were asked to select the score best representing the similarity between the predicted answer and the ground-truth answer. Annotators most frequently selected the score produced by a weight of $10$. On average, each annotator selected this weight $86\%$ of the times. For 73 samples on which the three annotator agreed, they selected this weight $96\%$ of the times.

\subsection{Setting the similarity threshold}
\label{app:threshold}

It is common practice in the field of Document VQA to set a threshold for NLS \cite{scene2019, docvqa2021, mpdocvqa2023, infographicvqa2022, anls*2024}. This is done to determine whether a match can be reasonably expected, or whether any similarity is coincidental (e.g. the NLS between ``dog'' and ``giraffe'' is larger than 0 as they share the letter ``g'', but the two are entirely different tokens). Following \citet{scene2019}, most studies have set the threshold to 0.5. Given that this was not justified by any validation study in \citet{scene2019}, we instead conducted our own tuning exercise using the validation dataset described in Appendix \ref{app:tuning}. Three human annotators performed a binary classification on 100 pairs of predicted answers and most similar spans from the corresponding documents. Each pair was tagged as a ``match'' or a ``mismatch'', indicating whether the predicted answer referred to the same span (perhaps with slight changes in spelling). The NLS value of 0.3 yielded the most optimal threshold for distinguishing between matching and mismatched pairs, predicting a mismatch with an F1 of $0.94$. 

\subsection{Calculating volatility}
\label{app:vol}

We use the standard definition of volatility as scaled standard deviation:

\begin{equation}
\text{vol}([x_1, \cdots, x_T]) = \text{std}([x_1, \cdots, x_T]) \sqrt{T}
\end{equation}

%% file: latex/appqtypes.tex
\section{Determining the types of questions in DocVQA}
\label{app:qtypes}

To determine the type of each question, we passed the following information to GPT-4o: 1) The document image. 2) The question. 3) The ground truth answer, as provided by the dataset. 4) A prompt, asking the model to determine the context from which the answer was extracted.

You can see an example prompt below:

\begin{prompt}[colupper=blue,fontupper=\bfseries\small]

\par Question: What is the extension number?
\par Answer: 5177

\par The above question was answered based on the document attached. What do you think best describes the context from which the answer was extracted? Select one of the below options. Simply return the correct option without any explanation.

\begin{enumerate}
\item Figure/Diagram
\item Form
\item Table/List
\item Layout
\item Free\_text
\item Image/Photo
\item Handwritten
\item Yes/No question
\item Other
\end{enumerate}
\end{prompt}

The experiment ran on September 7th, 2024. The agreement rate with the DocVQA validation set was 69.5\%.

%% file: latex/appreranking.tex
\section{Extended leaderboard analysis}
\label{app:reranking}

\begin{figure*}[htp]
\centering
\begin{minipage}{0.5\textwidth}
\includegraphics[width=\textwidth]{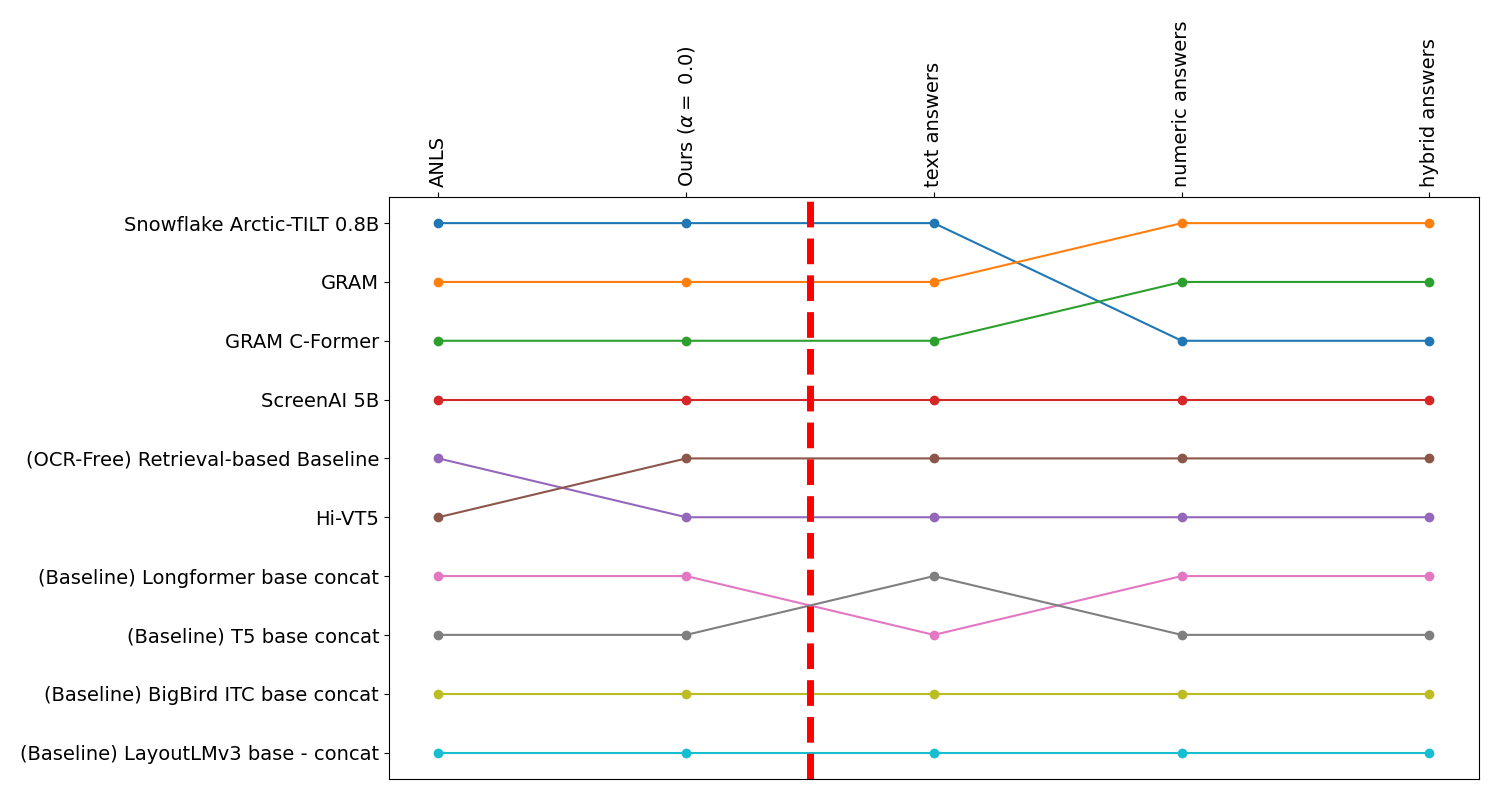}
\caption{MP-DocVQA leaderboard.}
\label{fig:mp_reranking}
\end{minipage}\hfill
\begin{minipage}{0.5\textwidth}
\includegraphics[width=\textwidth]{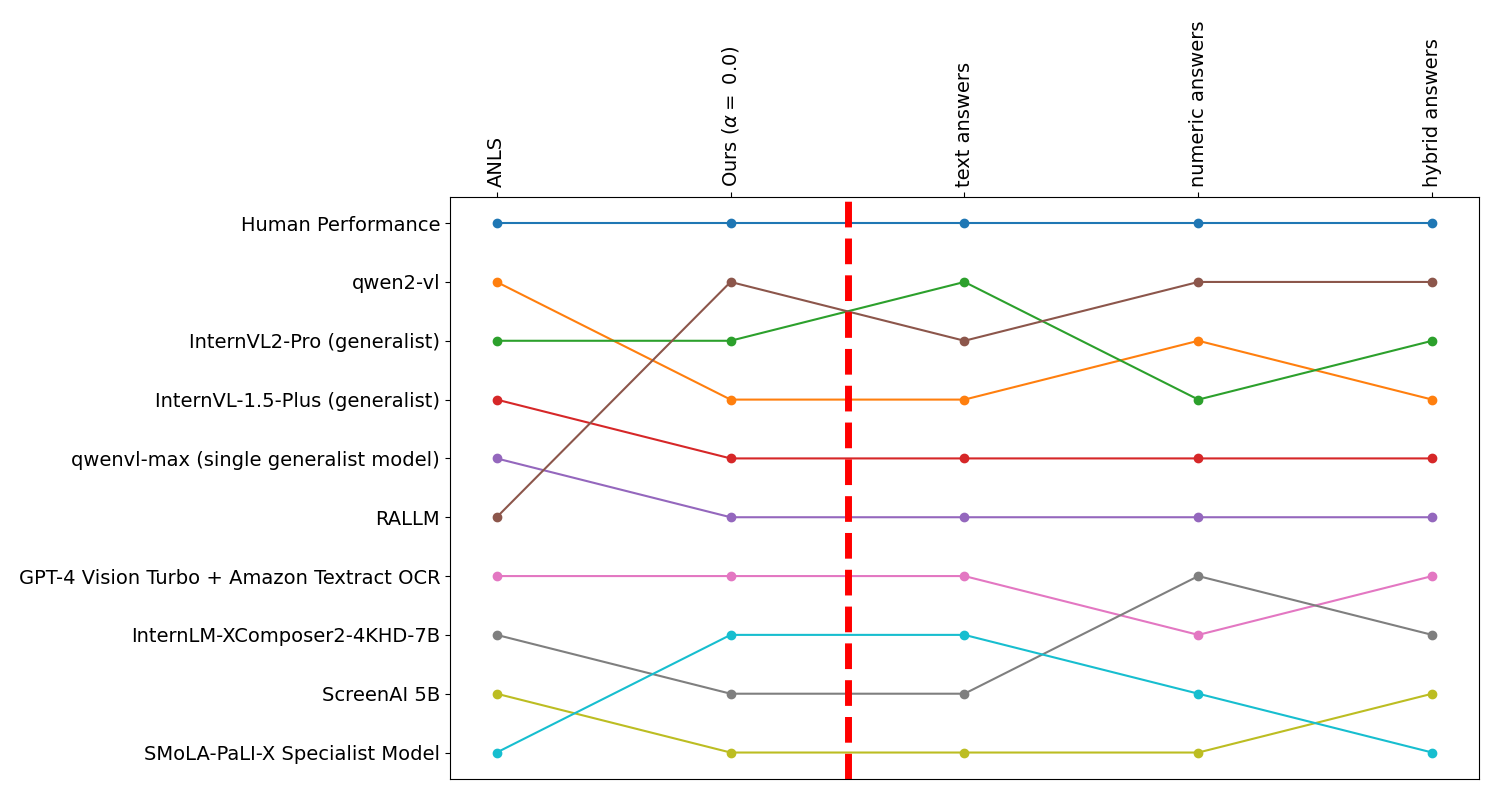}
\label{fig:info_reranking}
\caption{InfographicVQA leaderboard.}
\end{minipage}\par
\vskip\floatsep
\includegraphics[width=0.5\textwidth]{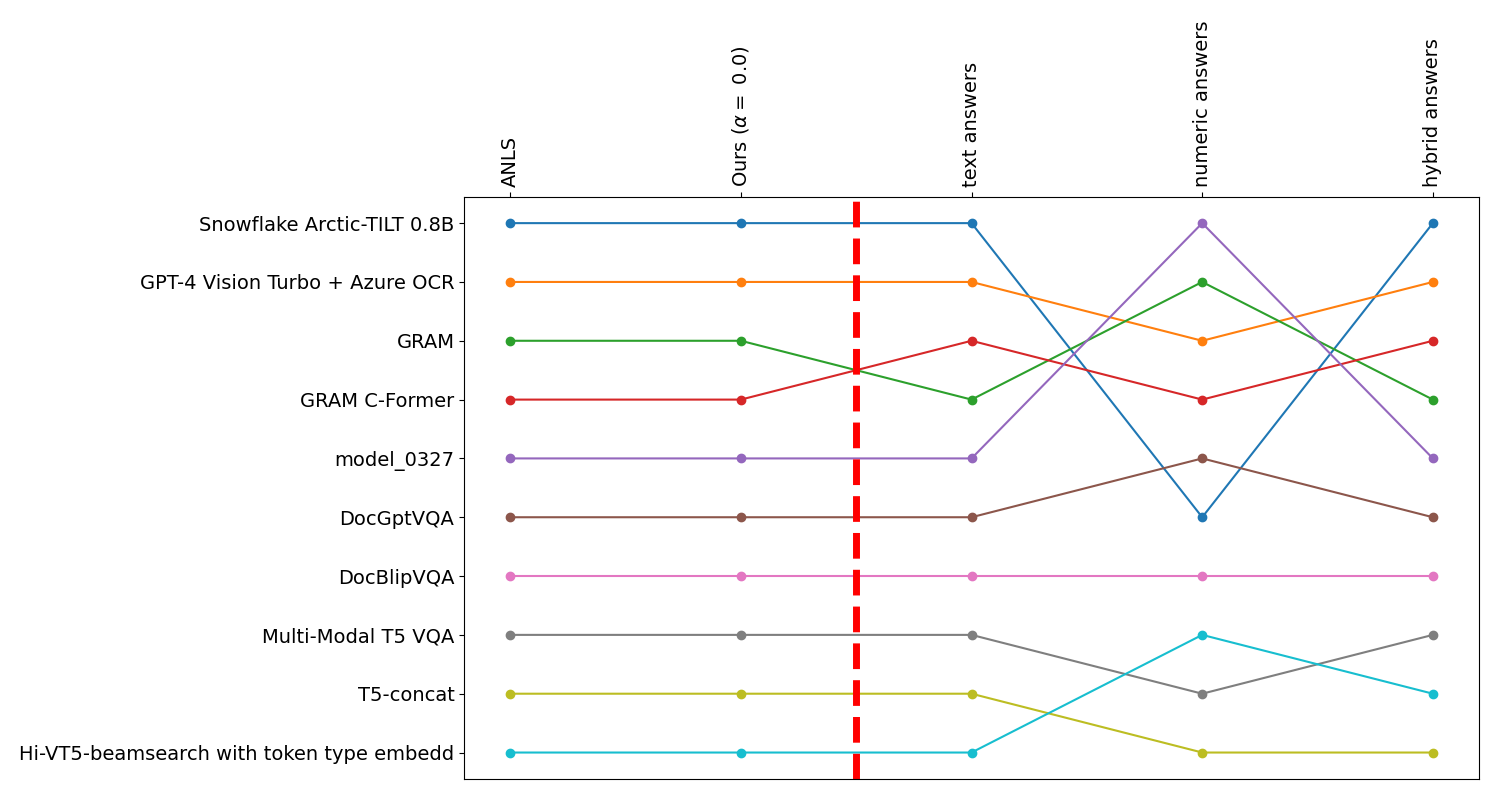}
\caption{DUDE leaderboard.}
\label{fig:dude_reranking}
\end{figure*}

Figures \ref{fig:mp_reranking} to \ref{fig:dude_reranking} show the reranking analysis for MP-DocVQA, InfographicVQA, and DUDE benchmarks, respectively. As with Figure \ref{fig:docvqa_reranking}, our composite score has been calculated with $\alpha=0.25$.

%% file: latex/appdocvqaqtypes.tex
\section{Extended question type analysis for DocVQA}
\label{app:docvqa}
\begin{figure*}
    \centering
    \begin{tabular}{cc}
        \begin{subfigure}[b]{0.5\textwidth}
            \centering
            \includegraphics[height=0.65\textwidth]{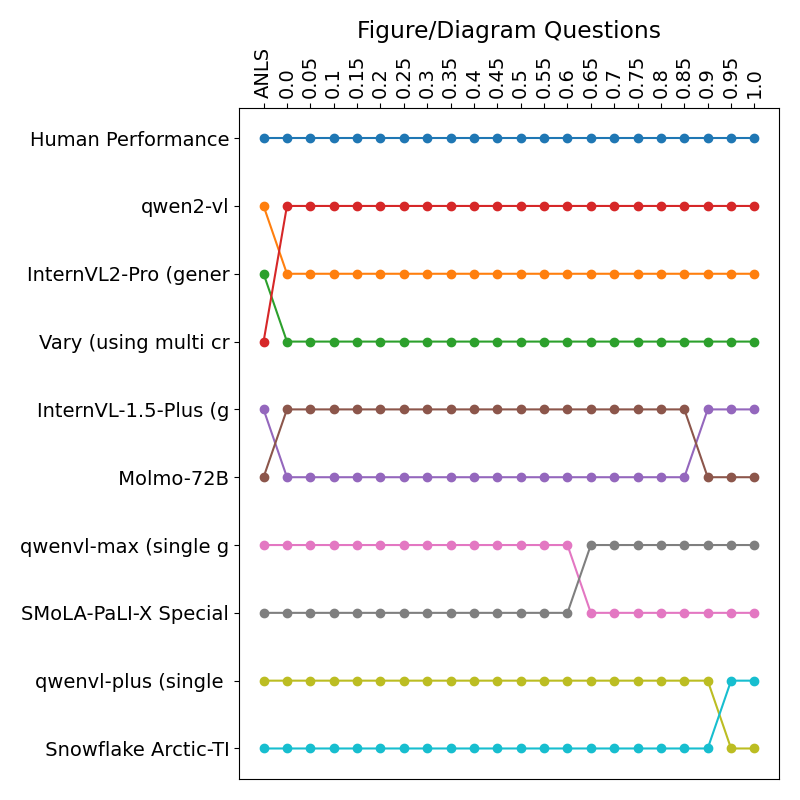}
            \caption{Figure/Diagram}
        \end{subfigure} &
        \begin{subfigure}[b]{0.5\textwidth}
            \centering
            \includegraphics[height=0.65\textwidth]{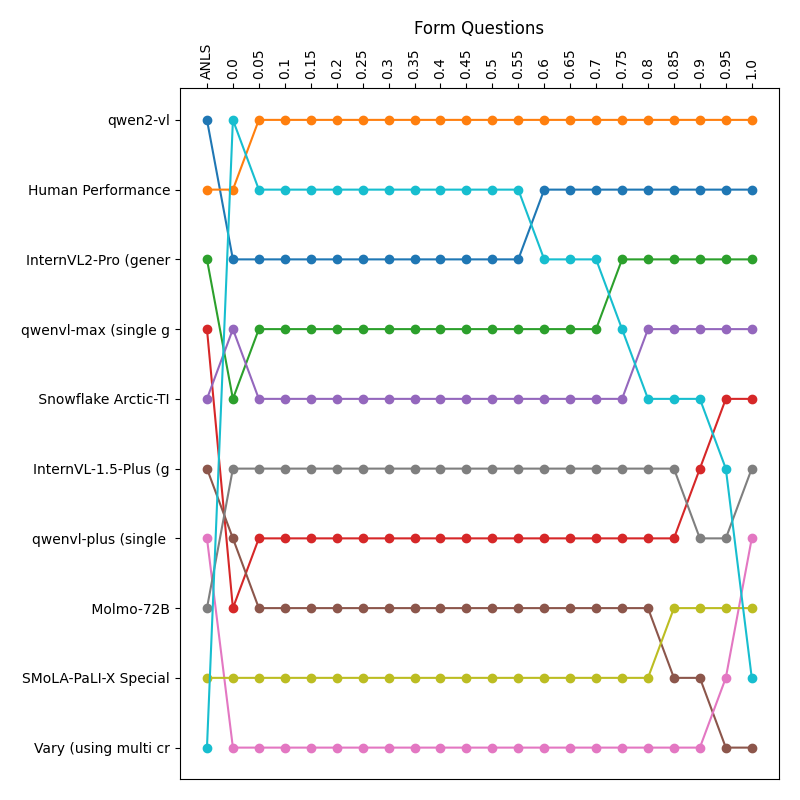}
            \caption{Form}
        \end{subfigure} \\
        \begin{subfigure}[b]{0.5\textwidth}
            \centering
            \includegraphics[height=0.65\textwidth]{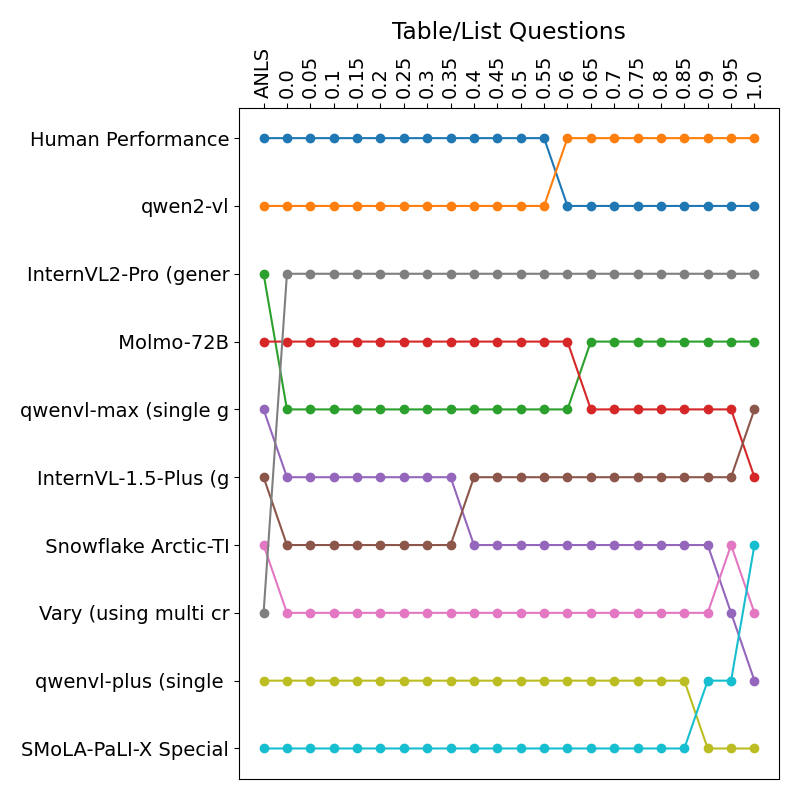}
            \caption{Table/List}
        \end{subfigure} &
        \begin{subfigure}[b]{0.5\textwidth}
            \centering
            \includegraphics[height=0.65\textwidth]{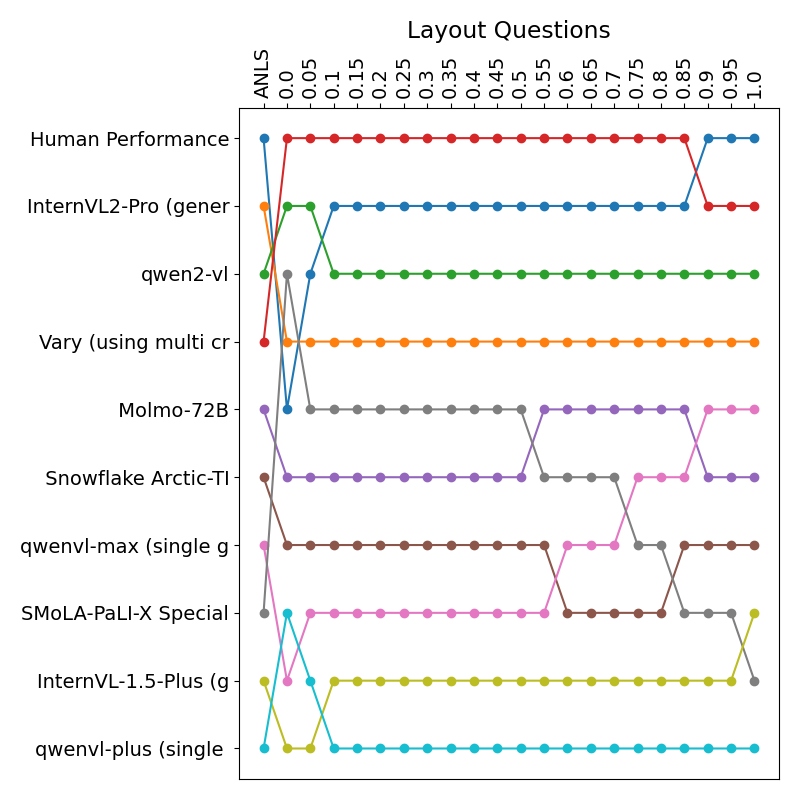}
            \caption{Layout}
        \end{subfigure} \\
        \begin{subfigure}[b]{0.5\textwidth}
            \centering
            \includegraphics[height=0.65\textwidth]{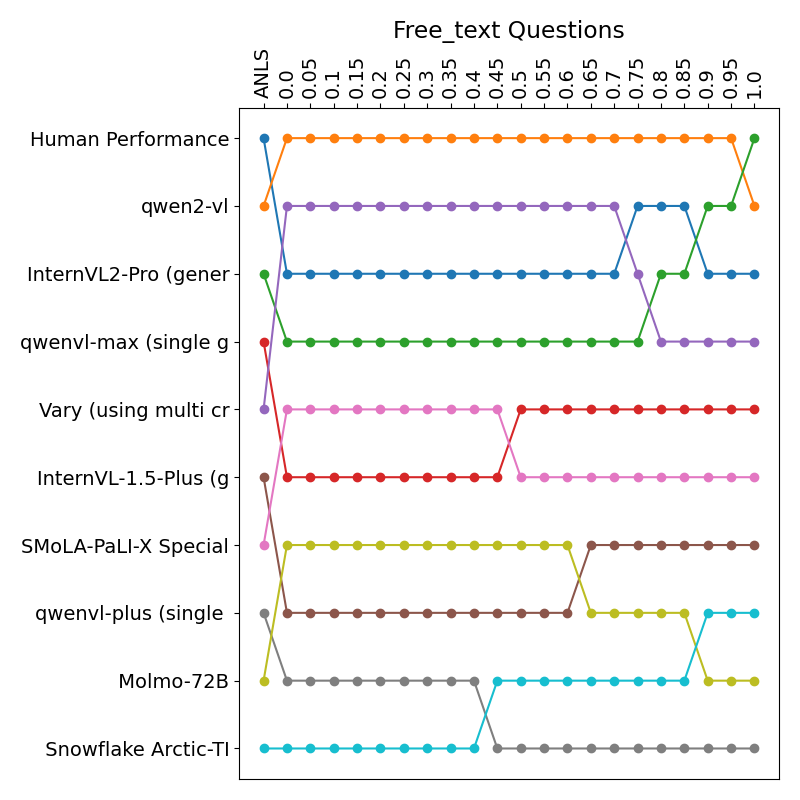}
            \caption{Free text}
        \end{subfigure} &
        \begin{subfigure}[b]{0.5\textwidth}
            \centering
            \includegraphics[height=0.65\textwidth]{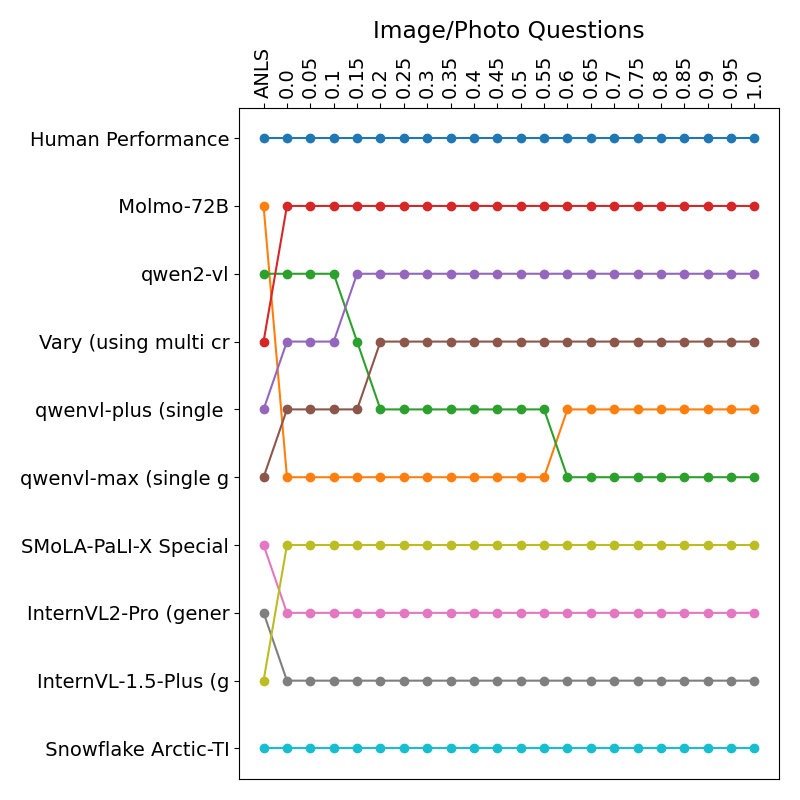}
            \caption{Image/Photo}
        \end{subfigure} \\
        \begin{subfigure}[b]{0.5\textwidth}
            \centering
            \includegraphics[height=0.65\textwidth]{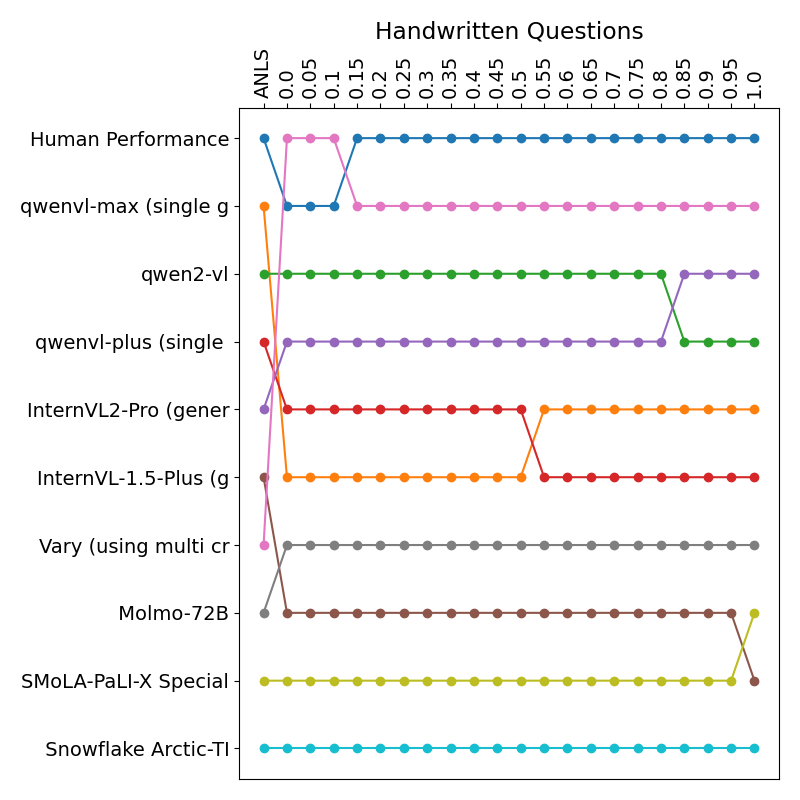}
            \caption{Handwritten}
        \end{subfigure} &
        \begin{subfigure}[b]{0.5\textwidth}
            \centering
            \includegraphics[height=0.65\textwidth]{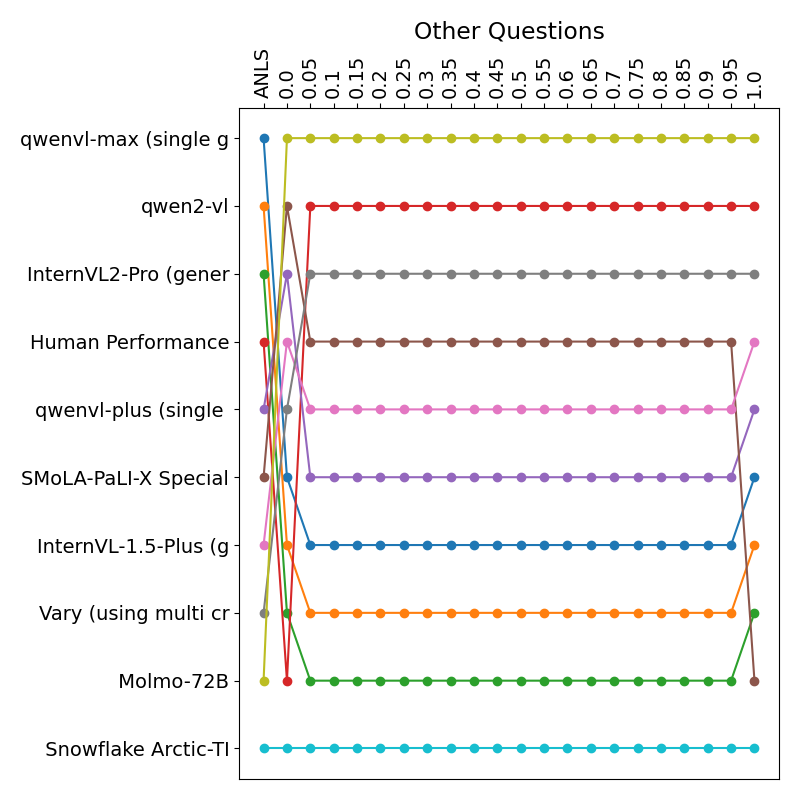}
            \caption{Other}
        \end{subfigure}
    \end{tabular}
    \caption{The impact of our score on the ranking of the top 10 models on the DocVQA benchmark, broken down by question type.}
    \label{fig:qtype_reranking}
\end{figure*}

Figure \ref{fig:qtype_reranking} shows how the top 10 models on the DocVQA leaderboard would be reranked if our score was used to evaluate them, broken down by question types.

%% file: latex/appdocvqaatypes.tex
\section{Answer type analysis for DocVQA}
\begin{figure*}
    \centering
    \begin{tabular}{cc}
        \begin{subfigure}[b]{0.45\textwidth}
            \centering
            \includegraphics[width=\textwidth]{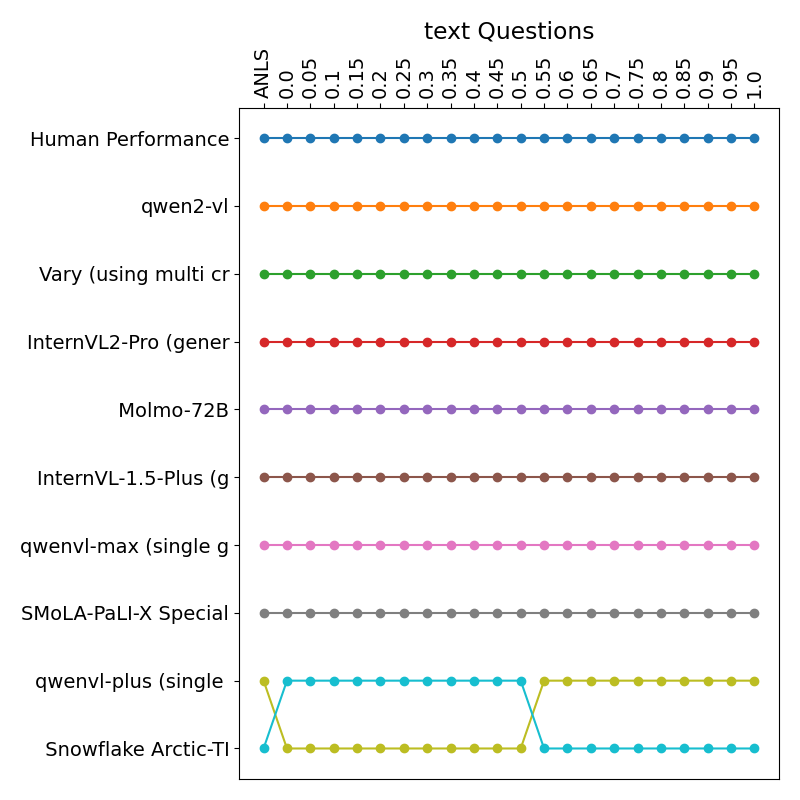}
            \caption{Textual}
        \end{subfigure} &
        \begin{subfigure}[b]{0.45\textwidth}
            \centering
            \includegraphics[width=\textwidth]{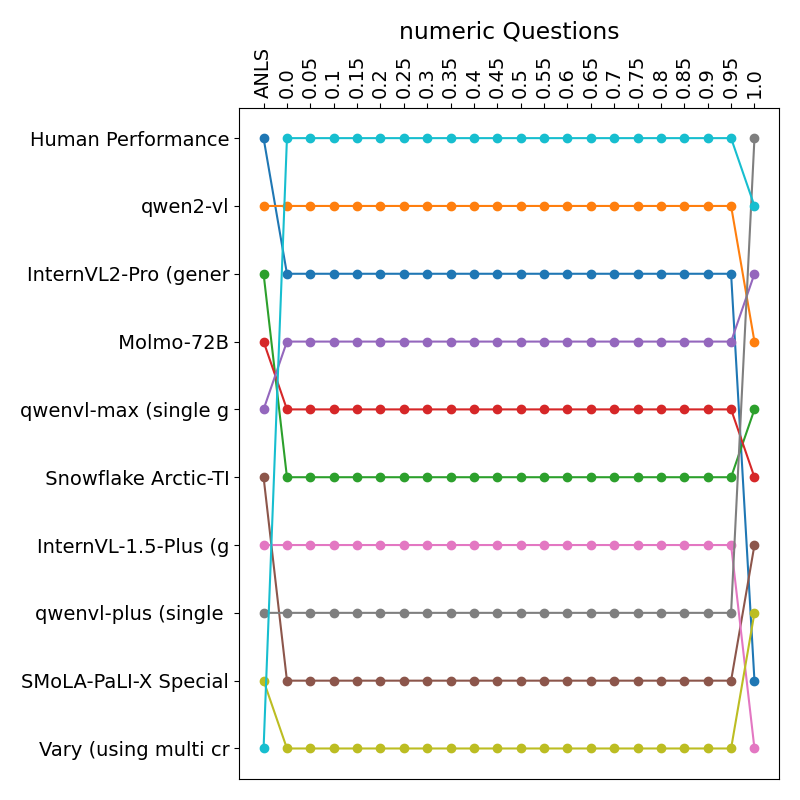}
            \caption{Numeric}
        \end{subfigure} \\
        \begin{subfigure}[b]{0.45\textwidth}
            \centering
            \includegraphics[width=\textwidth]{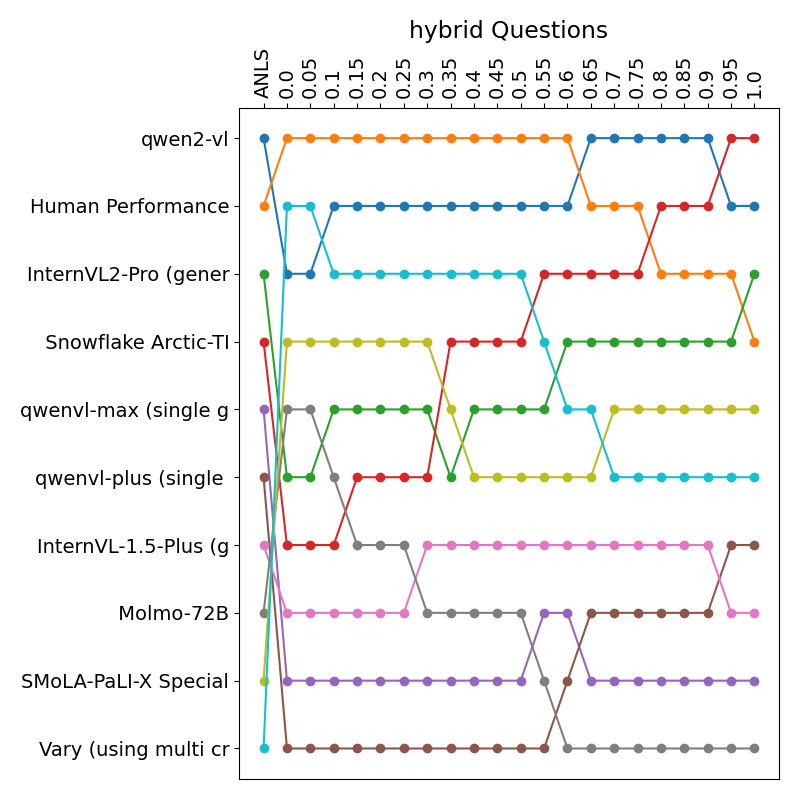}
            \caption{Hybrid}
        \end{subfigure} &
        \begin{subfigure}[b]{0.45\textwidth}
            \centering
            \includegraphics[width=\textwidth]{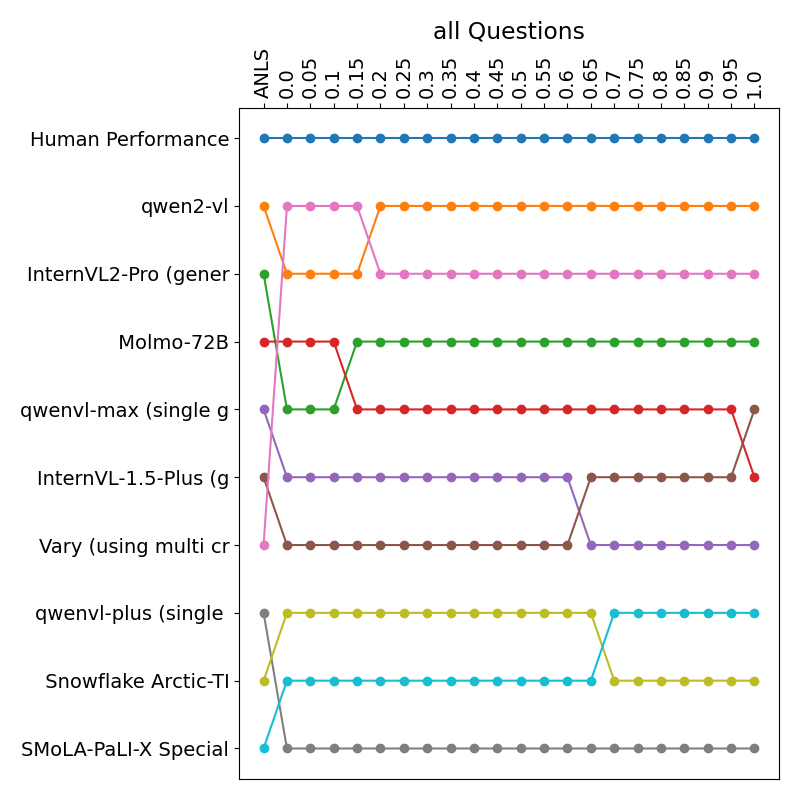}
            \caption{All}
        \end{subfigure}
    \end{tabular}
    \caption{The impact of our score on the ranking of the top 10 models on the DocVQA benchmark, broken down by answer type.}
    \label{fig:docvqa_atypes}
\end{figure*}

Figure \ref{fig:docvqa_atypes} shows how the top 10 models on the DocVQA leaderboard would be reranked if our score was used to evaluate them, broken down by answer types.

%% file: latex/appmpdocvqaatypes.tex
\section{Answer type analysis for MP-DocVQA}
\label{app:mpdocvqaatypes}
\begin{figure*}
    \centering
    \begin{tabular}{cc}
        \begin{subfigure}[b]{0.45\textwidth}
            \centering
            \includegraphics[width=\textwidth]{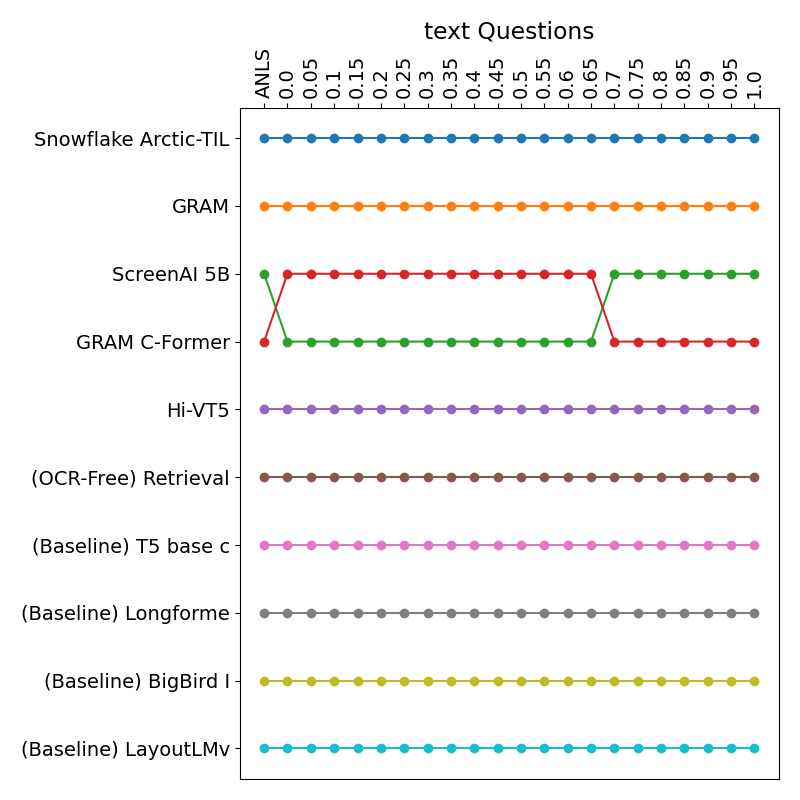}
            \caption{Textual}
        \end{subfigure} &
        \begin{subfigure}[b]{0.45\textwidth}
            \centering
            \includegraphics[width=\textwidth]{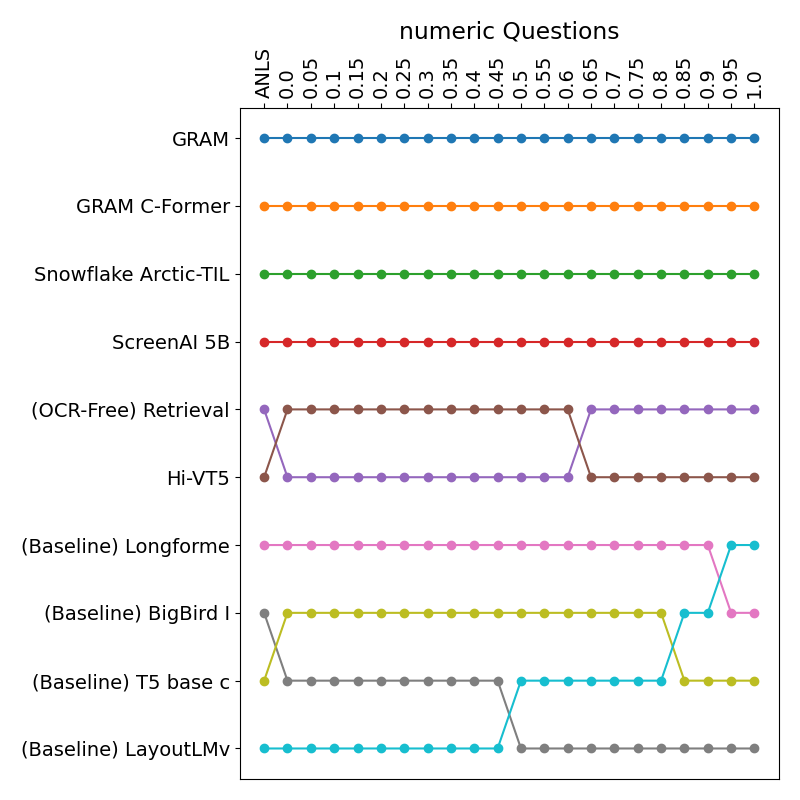}
            \caption{Numeric}
        \end{subfigure} \\
        \begin{subfigure}[b]{0.45\textwidth}
            \centering
            \includegraphics[width=\textwidth]{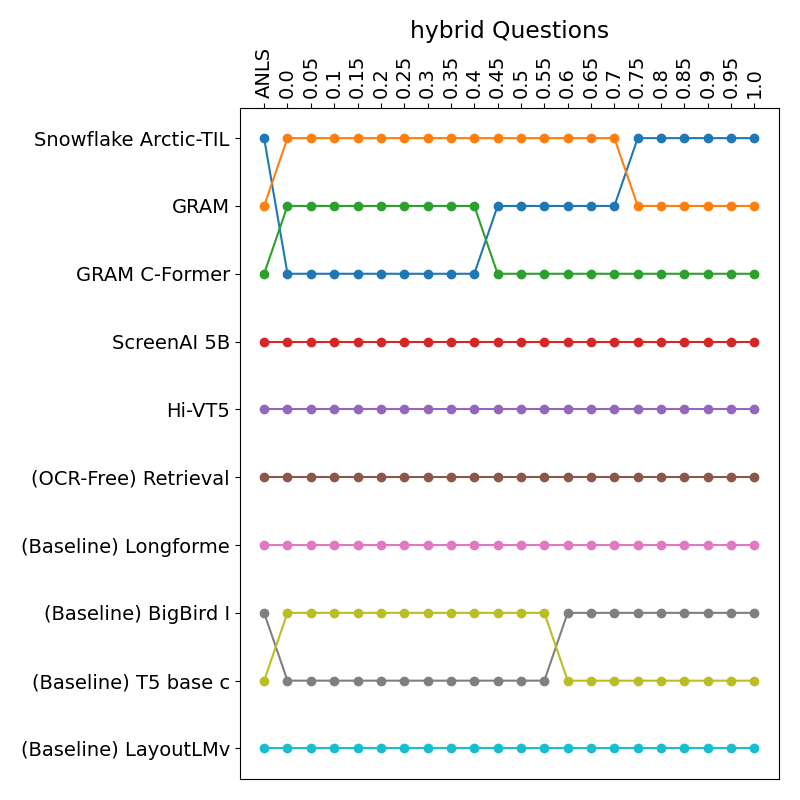}
            \caption{Hybrid}
        \end{subfigure} &
        \begin{subfigure}[b]{0.45\textwidth}
            \centering
            \includegraphics[width=\textwidth]{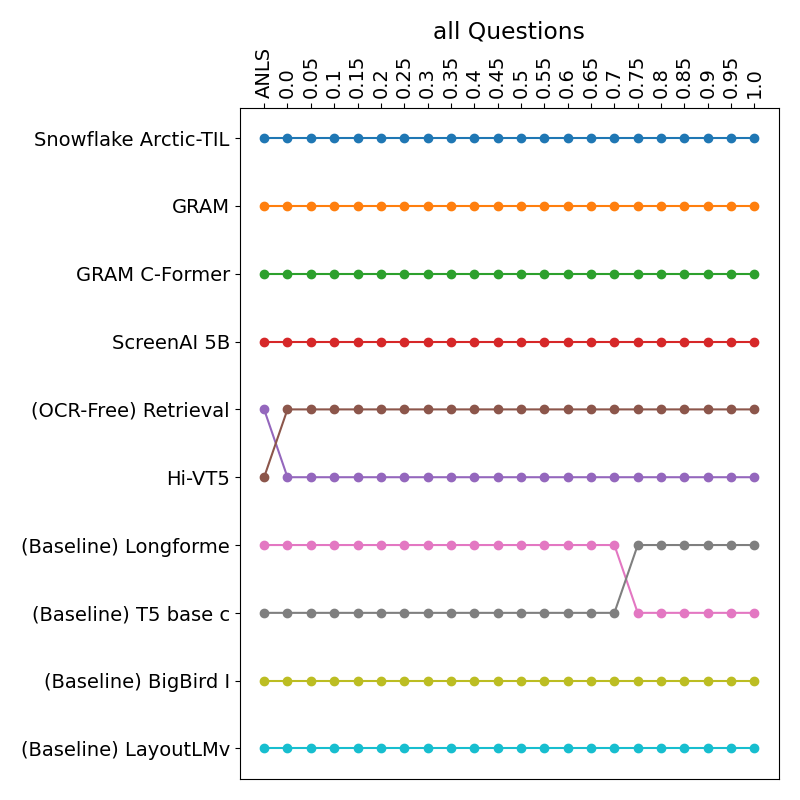}
            \caption{All}
        \end{subfigure}
    \end{tabular}
    \caption{The impact of our score on the ranking of the top 10 models on the MP-DocVQA benchmark, broken down by answer type.}
    \label{fig:mp_atypes}
\end{figure*}

Figure \ref{fig:mp_atypes} shows how the top 10 models on the MP-DocVQA leaderboard would be reranked if our score was used to evaluate them, broken down by answer types.

%% file: latex/appinfoatypes.tex
\section{Answer type analysis for InfographicVQA}
\label{app:infoatypes}
\begin{figure*}
    \centering
    \begin{tabular}{cc}
        \begin{subfigure}[b]{0.45\textwidth}
            \centering
            \includegraphics[width=\textwidth]{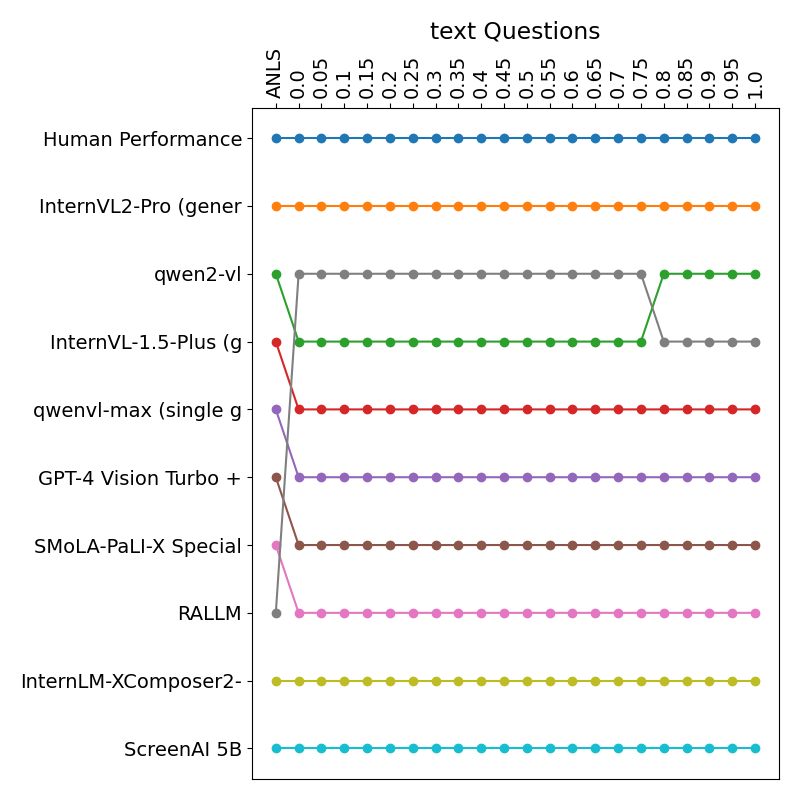}
            \caption{Textual}
        \end{subfigure} &
        \begin{subfigure}[b]{0.45\textwidth}
            \centering
            \includegraphics[width=\textwidth]{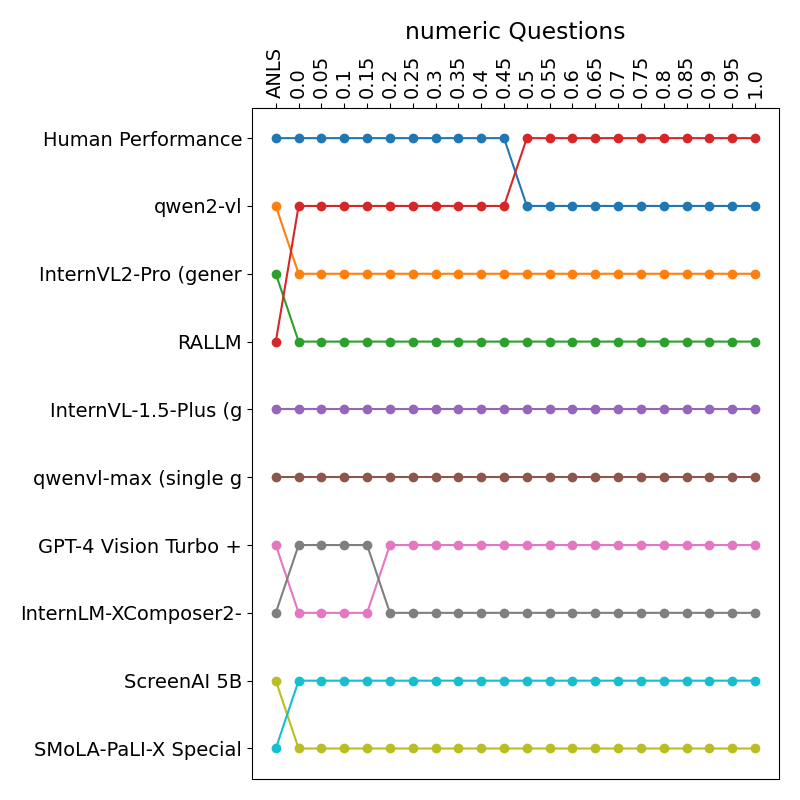}
            \caption{Numeric}
        \end{subfigure} \\
        \begin{subfigure}[b]{0.45\textwidth}
            \centering
            \includegraphics[width=\textwidth]{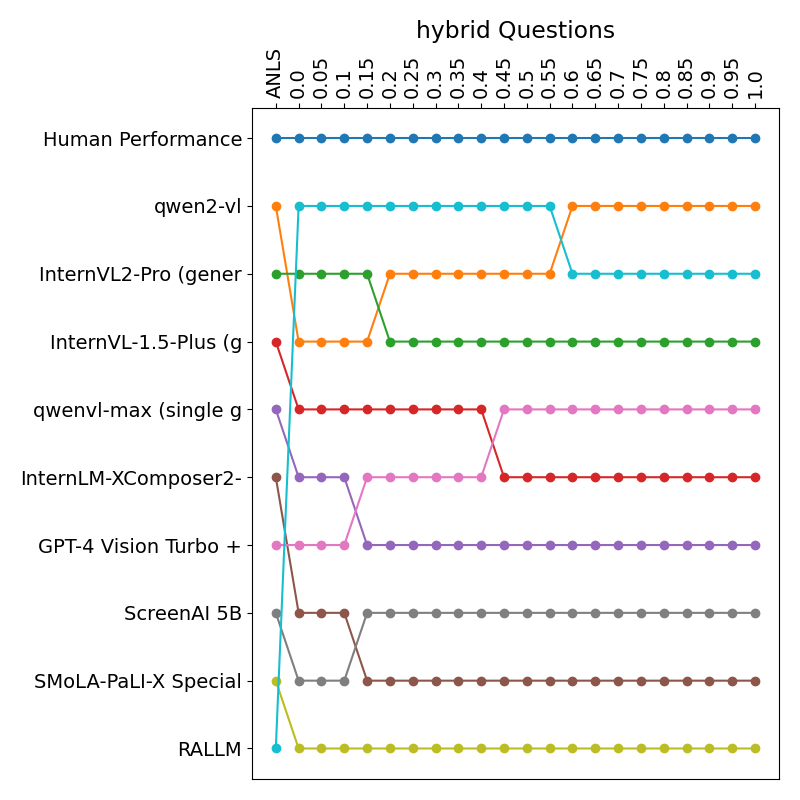}
            \caption{Hybrid}
        \end{subfigure} &
        \begin{subfigure}[b]{0.45\textwidth}
            \centering
            \includegraphics[width=\textwidth]{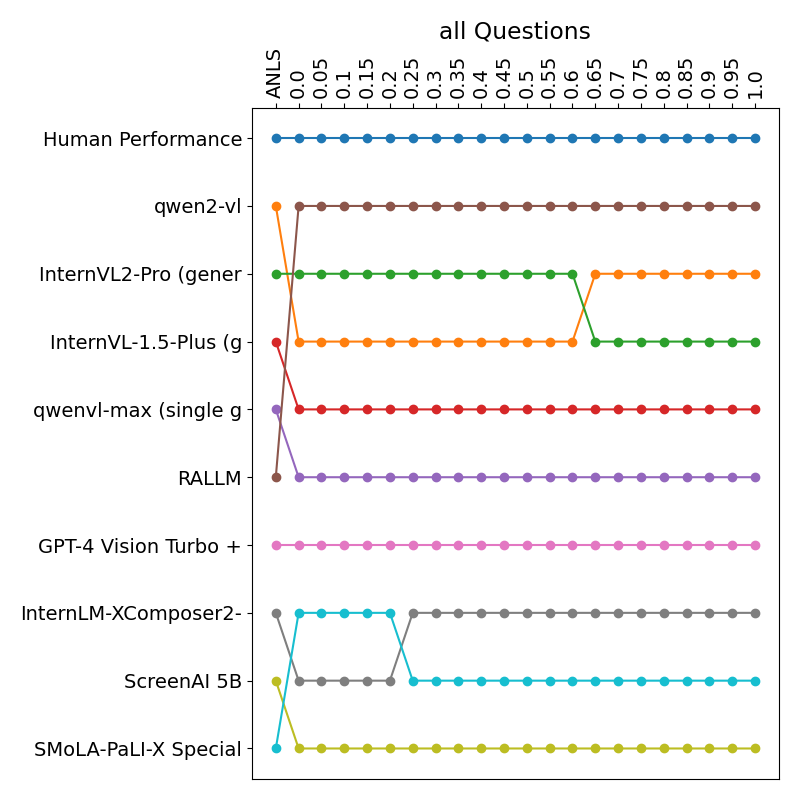}
            \caption{All}
        \end{subfigure}
    \end{tabular}
    \caption{The impact of our score on the ranking of the top 10 models on the InfographicVQA benchmark, broken down by answer type.}
    \label{fig:info_atypes}
\end{figure*}

Figure \ref{fig:info_atypes} shows how the top 10 models on the InfographicVQA leaderboard would be reranked if our score was used to evaluate them, broken down by answer types.

%% file: latex/appdudeatypes.tex
\section{Answer type analysis for DUDE}
\label{app:dudeatypes}
\begin{figure*}
    \centering
    \begin{tabular}{cc}
        \begin{subfigure}[b]{0.45\textwidth}
            \centering
            \includegraphics[width=\textwidth]{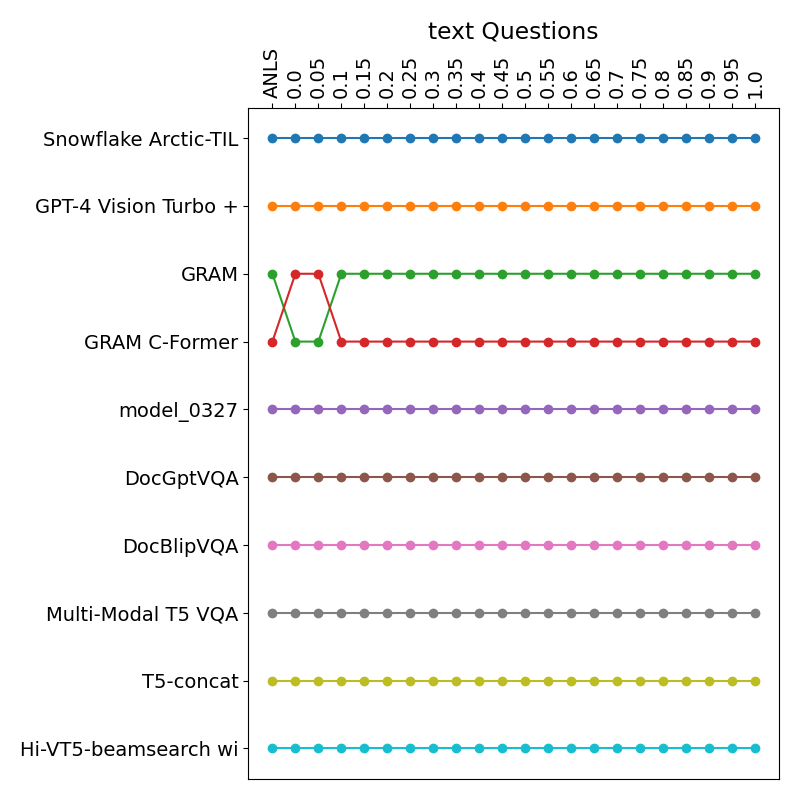}
            \caption{Textual}
        \end{subfigure} &
        \begin{subfigure}[b]{0.45\textwidth}
            \centering
            \includegraphics[width=\textwidth]{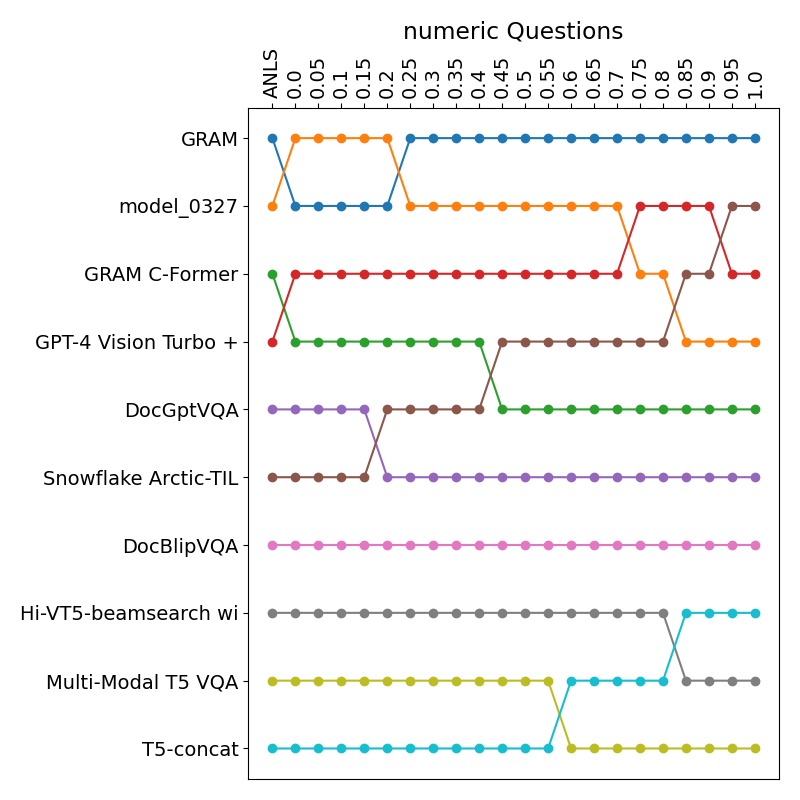}
            \caption{Numeric}
        \end{subfigure} \\
        \begin{subfigure}[b]{0.45\textwidth}
            \centering
            \includegraphics[width=\textwidth]{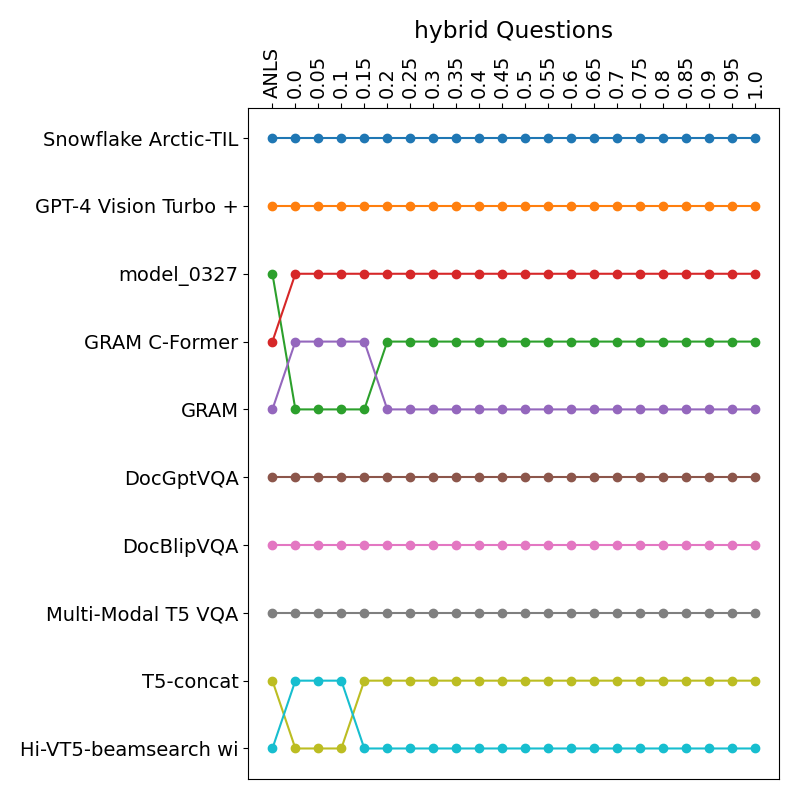}
            \caption{Hybrid}
        \end{subfigure} &
        \begin{subfigure}[b]{0.45\textwidth}
            \centering
            \includegraphics[width=\textwidth]{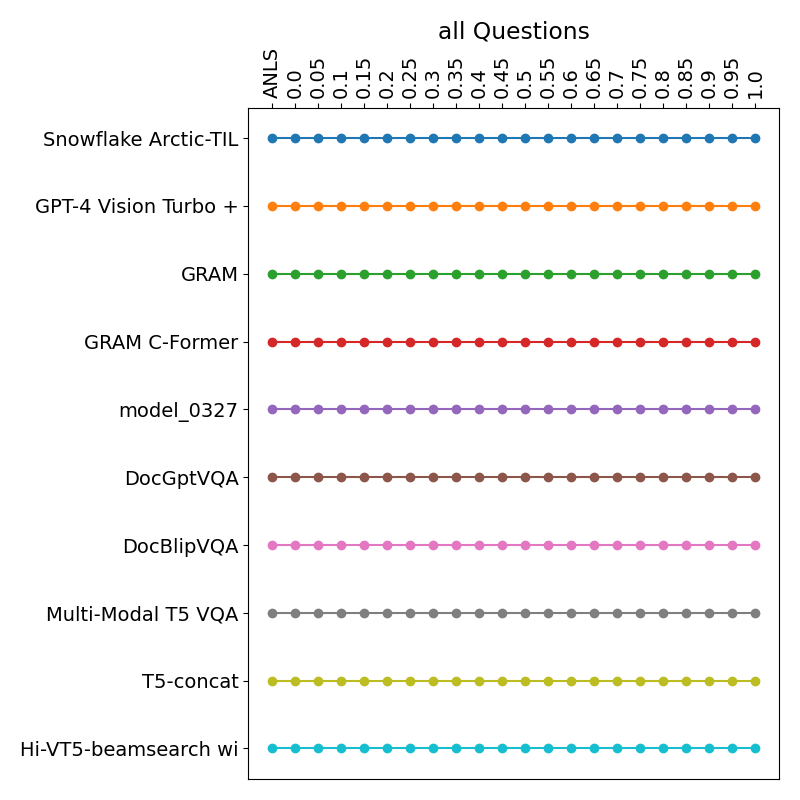}
            \caption{All}
        \end{subfigure}
    \end{tabular}
    \caption{The impact of our score on the ranking of the top 10 models on the DUDE benchmark, broken down by answer type.}
    \label{fig:dude_atypes}
\end{figure*}

Figure \ref{fig:dude_atypes} shows how the top 10 models on the DUDE leaderboard would be reranked if our score was used to evaluate them, broken down by answer types.

%% file: latex/appatypes.tex
\section{Correlation between answer types and original ranking}
\label{app:atypes}
\begin{figure*}
    \centering
    \begin{tabular}{cc}
        \begin{subfigure}[b]{0.45\textwidth}
            \centering
            \includegraphics[width=\textwidth]{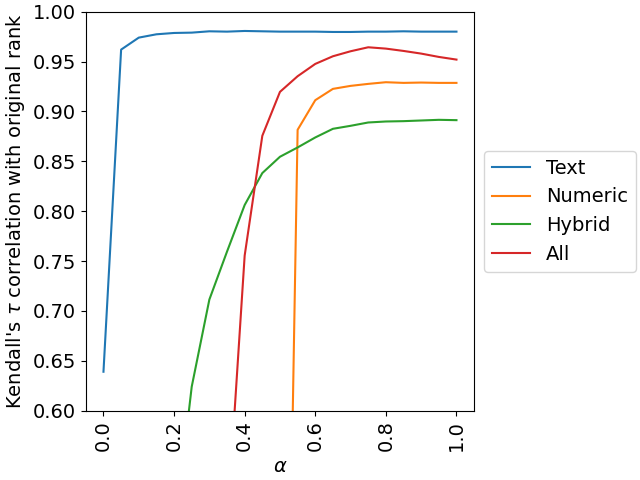}
            \caption{DocVQA}
        \end{subfigure} &
        \begin{subfigure}[b]{0.45\textwidth}
            \centering
            \includegraphics[width=\textwidth]{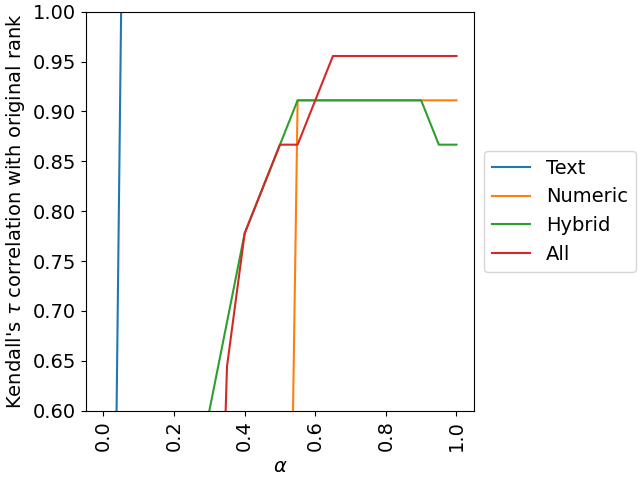}
            \caption{MP-DocVQA}
        \end{subfigure} \\
        \begin{subfigure}[b]{0.45\textwidth}
            \centering
            \includegraphics[width=\textwidth]{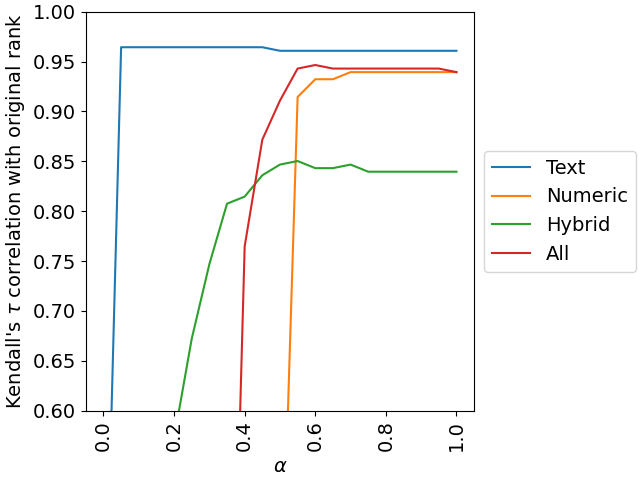}
            \caption{InfographicVQA}
        \end{subfigure} &
        \begin{subfigure}[b]{0.45\textwidth}
            \centering
            \includegraphics[width=\textwidth]{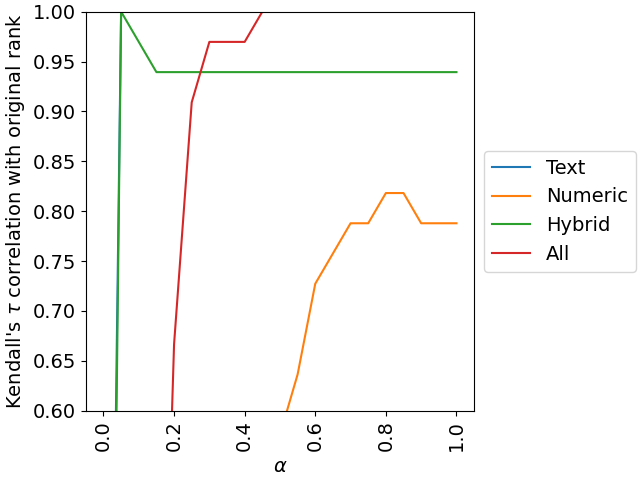}
            \caption{DUDE}
        \end{subfigure}
    \end{tabular}
    \caption{Kendall's $\tau$ correlation between different $\alpha$ settings and the original ranking of each benchmark, broken down by the type of answers.}
    \label{fig:all_atypes}
\end{figure*}

Figure \ref{fig:all_atypes} shows the correlation between the ranking of each leaderboard and the ranking produced by \metricName at various various for $\alpha$, broken down by the type of answer.

%% file: acl_latex.bbl
\begin{thebibliography}{25}
\providecommand{\natexlab}[1]{#1}

\bibitem[{gpt()}]{gpt4o}

\newblock Hello {GPT-4o}.
\newblock \url{https://openai.com/index/hello-gpt-4o/}.
\newblock Accessed: 2024-09-15.

\bibitem[{Alzahrani et~al.(2024)Alzahrani, Alyahya, Alnumay, AlRashed, Alsubaie, Almushayqih, Mirza, Alotaibi, Al-Twairesh, Alowisheq, Bari, and Khan}]{benchmarks2024}
Norah Alzahrani, Hisham Alyahya, Yazeed Alnumay, Sultan AlRashed, Shaykhah Alsubaie, Yousef Almushayqih, Faisal Mirza, Nouf Alotaibi, Nora Al-Twairesh, Areeb Alowisheq, M~Saiful Bari, and Haidar Khan. 2024.
\newblock \href {https://doi.org/10.18653/v1/2024.acl-long.744} {When benchmarks are targets: Revealing the sensitivity of large language model leaderboards}.
\newblock In \emph{Proceedings of the 62nd Annual Meeting of the Association for Computational Linguistics (Volume 1: Long Papers)}, pages 13787--13805, Bangkok, Thailand. Association for Computational Linguistics.

\bibitem[{Biten et~al.(2019)Biten, Tito, Mafla, Gomez, Rusinol, Mathew, Jawahar, Valveny, and Karatzas}]{scene2019}
Ali~Furkan Biten, Ruben Tito, Andres Mafla, Lluis Gomez, Mar{\c{c}}al Rusinol, Minesh Mathew, CV~Jawahar, Ernest Valveny, and Dimosthenis Karatzas. 2019.
\newblock \href {https://arxiv.org/abs/1907.00490} {{ICDAR} 2019 competition on scene text visual question answering}.
\newblock In \emph{2019 International Conference on Document Analysis and Recognition (ICDAR)}, pages 1563--1570. IEEE.

\bibitem[{Borchmann et~al.(2024)Borchmann, Pietruszka, Ja{\'s}kowski, Jurkiewicz, Halama, J{\'o}ziak, Garncarek, Liskowski, Szyndler, Gretkowski et~al.}]{tilt2024}
{\L}ukasz Borchmann, Micha{\l} Pietruszka, Wojciech Ja{\'s}kowski, Dawid Jurkiewicz, Piotr Halama, Pawe{\l} J{\'o}ziak, {\L}ukasz Garncarek, Pawe{\l} Liskowski, Karolina Szyndler, Andrzej Gretkowski, et~al. 2024.
\newblock \href {https://arxiv.org/abs/2408.04632} {{Arctic-TILT}. {Business} document understanding at sub-billion scale}.
\newblock \emph{arXiv preprint arXiv:2408.04632}.

\bibitem[{Chen et~al.(2024)Chen, Wang, Tian, Ye, Gao, Cui, Tong, Hu, Luo, Ma et~al.}]{internvl2024}
Zhe Chen, Weiyun Wang, Hao Tian, Shenglong Ye, Zhangwei Gao, Erfei Cui, Wenwen Tong, Kongzhi Hu, Jiapeng Luo, Zheng Ma, et~al. 2024.
\newblock \href {https://arxiv.org/abs/2404.16821} {How far are we to {GPT-4V}? closing the gap to commercial multimodal models with open-source suites}.
\newblock \emph{arXiv preprint arXiv:2404.16821}.

\bibitem[{Chi et~al.(2022)Chi, Fan, Ramadge, and Rudnicky}]{kerple2022}
Ta-Chung Chi, Ting-Han Fan, Peter~J. Ramadge, and Alexander~I. Rudnicky. 2022.
\newblock \href {https://api.semanticscholar.org/CorpusID:248965309} {Kerple: Kernelized relative positional embedding for length extrapolation}.
\newblock \emph{ArXiv}, abs/2205.09921.

\bibitem[{Deitke et~al.(2024)Deitke, Clark, Lee, Tripathi, Yang, Park, Salehi, Muennighoff, Lo, Soldaini, Lu, Anderson, Bransom, Ehsani, Ngo, Chen, Patel, Yatskar, Callison-Burch, Head, Hendrix, Bastani, VanderBilt, Lambert, Chou, Chheda, Sparks, Skjonsberg, Schmitz, Sarnat, Bischoff, Walsh, Newell, Wolters, Gupta, Borchardt, Groeneveld, Dumas, Nam, Lebrecht, Wittlif, Schoenick, Michel, Krishna, Weihs, Smith, Hajishirzi, Girshick, Farhadi, and Kembhavi}]{molmo2024}
Matt Deitke, Christopher Clark, Sangho Lee, Rohun Tripathi, Yue Yang, James Park, Reza Salehi, Niklas Muennighoff, Kyle Lo, Luca Soldaini, Jiasen Lu, Taira Anderson, Erin Bransom, Kiana Ehsani, Huong Ngo, YenSung Chen, Ajay Patel, Mark Yatskar, Chris Callison-Burch, Andrew Head, Rose Hendrix, Favyen Bastani, Eli VanderBilt, Nathan Lambert, Yvonne Chou, Arnavi Chheda, Jenna Sparks, Sam Skjonsberg, Michael Schmitz, Aaron Sarnat, Byron Bischoff, Pete Walsh, Chris Newell, Piper Wolters, Tanmay Gupta, Jon Borchardt, Dirk Groeneveld, Jen Dumas, Crystal Nam, Sophie Lebrecht, Caitlin Wittlif, Carissa Schoenick, Oscar Michel, Ranjay Krishna, Luca Weihs, Noah Smith, Hannaneh Hajishirzi, Ross Girshick, Ali Farhadi, and Aniruddha Kembhavi. 2024.
\newblock \href {https://molmo.allenai.org/paper.pdf} {Molmo and {PixMo}: Open weights and open data for state-of-the-art multimodal models}.

\bibitem[{Karatzas et~al.(2015)Karatzas, Gomez-Bigorda, Nicolaou, Ghosh, Bagdanov, Iwamura, Matas, Neumann, Chandrasekhar, Lu et~al.}]{localization2015}
Dimosthenis Karatzas, Lluis Gomez-Bigorda, Anguelos Nicolaou, Suman Ghosh, Andrew Bagdanov, Masakazu Iwamura, Jiri Matas, Lukas Neumann, Vijay~Ramaseshan Chandrasekhar, Shijian Lu, et~al. 2015.
\newblock \href {https://ieeexplore.ieee.org/document/7333942} {{ICDAR} 2015 competition on robust reading}.
\newblock In \emph{2015 13th international conference on document analysis and recognition (ICDAR)}, pages 1156--1160. IEEE.

\bibitem[{Lee et~al.(2021)Lee, Li, Wang, Wang, Fujii, Qin, Popat, and Pfister}]{rope2021}
Chen-Yu Lee, Chun-Liang Li, Chu Wang, Renshen Wang, Yasuhisa Fujii, Siyang Qin, Ashok Popat, and Tomas Pfister. 2021.
\newblock \href {https://doi.org/10.18653/v1/2021.acl-short.41} {{ROPE}: Reading order equivariant positional encoding for graph-based document information extraction}.
\newblock In \emph{Proceedings of the 59th Annual Meeting of the Association for Computational Linguistics and the 11th International Joint Conference on Natural Language Processing (Volume 2: Short Papers)}, pages 314--321, Online. Association for Computational Linguistics.

\bibitem[{Levenshtein(1966)}]{nls1966}
V~Levenshtein. 1966.
\newblock Binary codes capable of correcting deletions, insertions, and reversals.
\newblock In \emph{Soviet Physics-Doklady}, volume~10, pages 707--710.

\bibitem[{Mathew et~al.(2022)Mathew, Bagal, Tito, Karatzas, Valveny, and Jawahar}]{infographicvqa2022}
Minesh Mathew, Viraj Bagal, Rub\`en Tito, Dimosthenis Karatzas, Ernest Valveny, and C.V. Jawahar. 2022.
\newblock \href {https://openaccess.thecvf.com/content/WACV2022/papers/Mathew_InfographicVQA_WACV_2022_paper} {{InfographicVQA}}.
\newblock In \emph{Proceedings of the IEEE/CVF Winter Conference on Applications of Computer Vision (WACV)}, pages 1697--1706.

\bibitem[{Mathew et~al.(2021)Mathew, Karatzas, and Jawahar}]{docvqa2021}
Minesh Mathew, Dimosthenis Karatzas, and C.V. Jawahar. 2021.
\newblock \href {https://openaccess.thecvf.com/content/WACV2021/html/Mathew_DocVQA_A_Dataset_for_VQA_on_Document_Images_WACV_2021_paper.html} {{DocVQA}: {A Dataset} for {VQA} on {Document Images}}.
\newblock In \emph{Proceedings of the IEEE/CVF Winter Conference on Applications of Computer Vision (WACV)}, pages 2200--2209.

\bibitem[{Nourbakhsh et~al.(2024)Nourbakhsh, Shah, and Rose}]{position2024}
Armineh Nourbakhsh, Sameena Shah, and Carolyn Rose. 2024.
\newblock \href {https://doi.org/10.18653/v1/2024.findings-acl.870} {Towards a new research agenda for multimodal enterprise document understanding: What are we missing?}
\newblock In \emph{Findings of the Association for Computational Linguistics: ACL 2024}, pages 14610--14622, Bangkok, Thailand. Association for Computational Linguistics.

\bibitem[{Pakdaman~Naeini et~al.(2015)Pakdaman~Naeini, Cooper, and Hauskrecht}]{ece2015}
Mahdi Pakdaman~Naeini, Gregory Cooper, and Milos Hauskrecht. 2015.
\newblock \href {https://doi.org/10.1609/aaai.v29i1.9602} {Obtaining well calibrated probabilities using bayesian binning}.
\newblock \emph{Proceedings of the AAAI Conference on Artificial Intelligence}, 29(1).

\bibitem[{Peer et~al.(2024)Peer, Sch{\"o}pf, Nebendahl, Rietzler, and Stabinger}]{anls*2024}
David Peer, Philemon Sch{\"o}pf, Volckmar Nebendahl, Alexander Rietzler, and Sebastian Stabinger. 2024.
\newblock \href {https://arxiv.org/abs/2402.03848} {{ANLS*}--{A} universal document processing metric for generative large language models}.
\newblock \emph{arXiv preprint arXiv:2402.03848}.

\bibitem[{Qian et~al.(2024)Qian, Liu, Mao, Zhou, and Dou}]{grounddec2024}
Hongjin Qian, Zheng Liu, Kelong Mao, Yujia Zhou, and Zhicheng Dou. 2024.
\newblock \href {https://doi.org/10.18653/v1/2024.acl-long.71} {Grounding language model with chunking-free in-context retrieval}.
\newblock In \emph{Proceedings of the 62nd Annual Meeting of the Association for Computational Linguistics (Volume 1: Long Papers)}, pages 1298--1311, Bangkok, Thailand. Association for Computational Linguistics.

\bibitem[{Tito et~al.(2021)Tito, Karatzas, and Valveny}]{doccvqa2021}
Rub{\`e}n Tito, Dimosthenis Karatzas, and Ernest Valveny. 2021.
\newblock \href {https://link.springer.com/chapter/10.1007/978-3-030-86331-9_50} {Document collection visual question answering}.
\newblock In \emph{Document Analysis and Recognition--ICDAR 2021: 16th International Conference, Lausanne, Switzerland, September 5--10, 2021, Proceedings, Part II 16}, pages 778--792. Springer.

\bibitem[{Tito et~al.(2023)Tito, Karatzas, and Valveny}]{mpdocvqa2023}
Rubèn Tito, Dimosthenis Karatzas, and Ernest Valveny. 2023.
\newblock \href {https://doi.org/10.1016/j.patcog.2023.109834} {Hierarchical multimodal transformers for multipage docvqa}.
\newblock \emph{Pattern Recognition}, 144:109834.

\bibitem[{Van~Landeghem et~al.(2023)Van~Landeghem, Tito, Borchmann, Pietruszka, Joziak, Powalski, Jurkiewicz, Coustaty, Anckaert, Valveny, Blaschko, Moens, and Stanislawek}]{dude2023}
Jordy Van~Landeghem, Rub\`en Tito, {\L}ukasz Borchmann, Micha{\l} Pietruszka, Pawel Joziak, Rafal Powalski, Dawid Jurkiewicz, Mickael Coustaty, Bertrand Anckaert, Ernest Valveny, Matthew Blaschko, Sien Moens, and Tomasz Stanislawek. 2023.
\newblock \href {https://openaccess.thecvf.com/content/ICCV2023/html/Van_Landeghem_Document_Understanding_Dataset_and_Evaluation_DUDE_ICCV_2023_paper.html} {{Document Understanding Dataset and Evaluation (DUDE)}}.
\newblock In \emph{Proceedings of the IEEE/CVF International Conference on Computer Vision (ICCV)}, pages 19528--19540.

\bibitem[{Wang et~al.(2023)Wang, Ma, Nourbakhsh, Gu, and Shah}]{docgraphlm2023}
Dongsheng Wang, Zhiqiang Ma, Armineh Nourbakhsh, Kang Gu, and Sameena Shah. 2023.
\newblock \href {https://doi.org/10.1145/3539618.3591975} {{DocGraphLM}: Documental graph language model for information extraction}.
\newblock In \emph{Proceedings of the 46th International ACM SIGIR Conference on Research and Development in Information Retrieval}, SIGIR '23, page 1944–1948, New York, NY, USA. Association for Computing Machinery.

\bibitem[{Wang et~al.(2024)Wang, Bai, Tan, Wang, Fan, Bai, Chen, Liu, Wang, Ge et~al.}]{qwen2vl2024}
Peng Wang, Shuai Bai, Sinan Tan, Shijie Wang, Zhihao Fan, Jinze Bai, Keqin Chen, Xuejing Liu, Jialin Wang, Wenbin Ge, et~al. 2024.
\newblock \href {https://arxiv.org/abs/2409.12191} {{Qwen2-VL}: Enhancing vision-language model's perception of the world at any resolution}.
\newblock \emph{arXiv preprint arXiv:2409.12191}.

\bibitem[{Wu et~al.(2024)Wu, Hu, Wang, Pang, and Soricut}]{smolapalix2024}
Jialin Wu, Xia Hu, Yaqing Wang, Bo~Pang, and Radu Soricut. 2024.
\newblock \href {https://openaccess.thecvf.com/content/CVPR2024/papers/Wu_Omni-SMoLA_Boosting_Generalist_Multimodal_Models_with_Soft_Mixture_of_Low-rank_CVPR_2024_paper.pdf} {{Omni-SMoLA}: Boosting generalist multimodal models with soft mixture of low-rank experts}.
\newblock In \emph{Proceedings of the IEEE/CVF Conference on Computer Vision and Pattern Recognition}, pages 14205--14215.

\bibitem[{Zhang et~al.(2023)Zhang, Guo, Tu, Chen, Tang, Zhu, Zhang, and Gui}]{order2023}
Chong Zhang, Ya~Guo, Yi~Tu, Huan Chen, Jinyang Tang, Huijia Zhu, Qi~Zhang, and Tao Gui. 2023.
\newblock \href {https://arxiv.org/abs/2310.11016} {Reading order matters: Information extraction from visually-rich documents by token path prediction}.
\newblock \emph{arXiv preprint arXiv:2310.11016}.

\bibitem[{Zmigrod et~al.(2024{\natexlab{a}})Zmigrod, Ma, Nourbakhsh, and Shah}]{treeform2024}
Ran Zmigrod, Zhiqiang Ma, Armineh Nourbakhsh, and Sameena Shah. 2024{\natexlab{a}}.
\newblock \href {https://aclanthology.org/2024.law-1.1} {{T}ree{F}orm: End-to-end annotation and evaluation for form document parsing}.
\newblock In \emph{Proceedings of The 18th Linguistic Annotation Workshop (LAW-XVIII)}, pages 1--11, St. Julians, Malta. Association for Computational Linguistics.

\bibitem[{Zmigrod et~al.(2024{\natexlab{b}})Zmigrod, Shetty, Sibue, Ma, Nourbakhsh, Liu, and Veloso}]{k2q2024}
Ran Zmigrod, Pranav Shetty, Mathieu Sibue, Zhiqiang Ma, Armineh Nourbakhsh, Xiaomo Liu, and Manuela Veloso. 2024{\natexlab{b}}.
\newblock \href {https://doi.org/10.18653/v1/2024.findings-emnlp.770} {{\textquotedblleft}{What} is the value of \{templates\}?{\textquotedblright} {Rethinking} document information extraction datasets for {LLM}s}.
\newblock In \emph{Findings of the Association for Computational Linguistics: EMNLP 2024}, pages 13162--13185, Miami, Florida, USA. Association for Computational Linguistics.

\end{thebibliography}
